\documentclass{bmvc2k}

\title{MxT: Mamba x Transformer for \\Image Inpainting}
\addauthor{Shuang Chen}{shuang.chen@durham.ac.uk}{1}
\addauthor{Amir Atapour-Abarghouei}{amir.atapour-abarghouei@durham.ac.uk}{1}
\addauthor{Haozheng Zhang}{haozheng.zhang@durham.ac.uk}{1}
\addauthor{Hubert P. H. Shum$^{\dag}$}{hubert.shum@durham.ac.uk}{1}
\addinstitution{
 The department of Computer Science\\
 Durham University\\
 Durham, UK\\
 \vspace{0.3cm}
 $^{\dag}$Corresponding Author: Hubert P. H. Shum
}

\runninghead{M$\times$T}{Mamba $\times$ Transformer for Image Inpainting}

\usepackage{amsfonts}
\usepackage{tabu}
\usepackage{multirow}
\usepackage{multicol}
\usepackage{makecell}

\newcommand{\name}[1]{M$\times$T}
\newcommand\T{\rule{0pt}{2.9ex}}       
\newcommand\B{\rule[-1.2ex]{0pt}{0pt}} 
\usepackage[export]{adjustbox}
\usepackage{adjustbox}
\begin{document}

\maketitle

\begin{abstract}
Image inpainting, or image completion, is a crucial task in computer vision that aims to restore missing or damaged regions of images with semantically coherent content. This technique requires a precise balance of local texture replication and global contextual understanding to ensure the restored image integrates seamlessly with its surroundings. Traditional methods using Convolutional Neural Networks (CNNs) are effective at capturing local patterns but often struggle with broader contextual relationships due to the limited receptive fields. Recent advancements have incorporated transformers, leveraging their ability to understand global interactions. However, these methods face computational inefficiencies and struggle to maintain fine-grained details. To overcome these challenges, we introduce \name{} composed of the proposed Hybrid Module (HM), which combines Mamba with the transformer in a synergistic manner. Mamba is adept at efficiently processing long sequences with linear computational costs, making it an ideal complement to the transformer for handling long-scale data interactions. Our HM facilitates dual-level interaction learning at both pixel and patch levels, greatly enhancing the model to reconstruct images with high quality and contextual accuracy. We evaluate \name{} on the widely-used CelebA-HQ and Places2-standard datasets, where it consistently outperformed existing state-of-the-art methods. The code will be released: {\url{https://github.com/ChrisChen1023/MxT}}.

\end{abstract}

\section{Introduction}
Image inpainting, as known as image completion, aims at restoring missing or damaged parts of images with semantically plausible context. This demands accurate modeling of both global and local information within the corrupted image, which is crucial as the global and local interaction maintains the coherence of both the content and style of the missing areas, ensuring seamless integration with the surrounding image regions~\cite{yu2018generative}.

Convolutional Neural Networks (CNNs) have been employed for image inpainting, capitalizing on their ability to capture local patterns and textures. However, CNNs-based methods are inherently limited by the slow-grown receptive field, which limits the ability to grasp broader image context~\cite{li2022mat, zamir2022restormer}. To solve this issue, recent advancements~\cite{li2022mat, chen2024hint} have seen the integration of transformer or self-attention into image inpainting, leveraging their capability to capture global correlations across entire images. However, transformer-based methods are often constrained by quadratic computational complexity, prompting most methods to process images in smaller patches to reduce the spatial dimension~\cite{wan2021high, zheng2022bridging}, to learn the interaction in patch-level. This patch-based approach hinders the learning of fine-grained details, often resulting in artifacts in the generated images.

Mamba~\cite{gu2023mamba}, merging from the domain of long-sequence modeling, offers promising advantages for handling long sequential data and capturing long-range dependency efficiently, all at a linear computational cost. This capability makes Mamba particularly suitable for globally learning interactions at the pixel level, thus complementing transformers by adding detailed context.

We observe that, Mamba and transformer exhibit complementary strengths: Mamba is good at learning long-range pixel-wise dependency, which is computationally expensive for the transformer. Conversely, transformer is good at capturing global interactions between localized patches, such spatial awareness is an area that Mamba lacks due to it being designed for sequence modelling.

In this paper, we introduce \name{}, consisting of proposed Hybrid Modules that synergistically combine the strengths of both transformer and Mamba. This novel approach allows for dual-level interaction learning from the patch level and pixel level. Our comparative experiments demonstrate that our \name{} outperforms existing state-of-the-art methods on two wildly used datasets, CelebA-HQ and Places2. We summarize our contributions as: 1) We propose \name{} to introduce Mamba combined with transformer for image inpainting. 2) We design a novel Hybrid Module to capture the feature interaction at both the pixel level and patch level. 3) Our \name{} overall suppress the state-of-the-art methods on both CelebA-HQ and Places2 datasets. 4) \name{} is able to adapt to high-resolution images with only training on low-resolution data.

\label{sec:intro}

\section{Related Work}
\subsection{Image Inpainting} Image inpainting is an ill-posed low-level vision task that aims to infer the missing regions of the image via the undamaged pixels. Conventional works employ diffusion-based approaches to inpaint the missing regions by deriving by neighbouring visible pixels~\cite{sridevi2019image}, or filling the missing areas by using the good match patches from the background or external sources such as the depth information~\cite{atapour2016back,barnes2009patchmatch}. Although these methods can effectively complete small missing regions, they face challenges in precisely reconstructing more complex scenes due to limited global understanding of the image. Recently, deep learning based image inpainting studies propose to use CNN-based encoder-decoder architecture~\cite{yan2018shift, yu2020region,suin2021distillation} or CNN-based Generative Adversarial Networks (GANs)~\cite{xu2014deep,yu2018generative,iizuka2017globally,yi2020contextual,peng2021generating,sargsyan2023mi}, which significantly improves the visual plausibility and diversity of inpainted images. However, the limited convolutional receptive field hinders the model's learning of long-range dependencies, which motivates studies on expanding the receptive field by applying the frequency convolution techniques~\cite{suvorov2022resolution,chu2023rethinking} or developing transformer-based models~\cite{chen2024hint,li2022mat,chen2021pre,liang2021swinir,chang2023design}. Nonetheless, practical training and application of the transformer-based models are still constrained by the exponential complexity of self-attention calculations. In particular, it is still challenging to use pixel-level self-attention to achieve relatively high-resolution image inpainting. To this end, we turn our eyes to the field of selective SSM (i.e., Mamba~\cite{gu2023mamba}) for achieving image long-range pixel-wise dependency learning with robust spatial awareness.

\subsection{SSMs in Computer Vision} 
Recently, State Space Models (SSMs) have demonstrated promising advantages of long sequence modeling and linear-time complexity in Natural Language Processing (NLP)~\cite{gu2021efficiently}. This work specifically tackles the problem of vanishing gradients in SSMs when solving the exponential function by the linear first-order Ordinary Differential Equations~\cite{gu2020hippo}. Building on the rigorous theoretical proofs of the HiPPO framework that enables SSMs to capture long-range dependencies, Gu et al.~\cite{gu2023mamba} further introduce a data-dependent selective structure SSM (i.e., Mamba) to significantly improve the computational efficiencies in conventional SSMs. Inspired by pioneering SSMs, vision-specific adaptations of the Mamba architecture, such as Vision Mamba~\cite{zhu2024vision} and V-Mamba~\cite{liu2024vmamba}, propose visual SSMs designs for computer vision tasks including image classification and object detection~\cite{liu2024vmamba,zhu2024vision}. However, their performance is still behind of the state-of-the-art transformer-based models like SpectFormer~\cite{patro2023spectformer}, SVT~\cite{patro2024scattering}, and WaveViT~\cite{yao2022wave}. U-Mamba~\cite{ma2024u} effectively extends the capabilities of Mamba for biomedical image segmentation by proposing a hybrid CNN-SSM block. However, these studies ineffectively leverage the capabilities of Mamba in image long-range pixel-level dependency learning and overlook the critical spatial awareness during model designs.

\section{Preliminary}
\label{pre_ssm}
\subsection{State Space Models and Mamba} 
The State Space Model (SSM) is generally known as a linear time-invariant system that maps a 1-dimensional input sequence $x(t)\in \mathbb{R}$ to a response $y(t)\in \mathbb{R}$ via a hidden latent state $h(t)\in \mathbb{R}^N$ (Eq.\ref{eq1}). For efficient linear-complexity deep learning model training, the structured SSMs employ a zero-order hold discretization rule (Eq.\ref{eq2}) to transform the continues parameter $(\Delta,\textbf{\textit{A}},\textbf{\textit{B}})$ to discrete parameters $(\overline{\textbf{\textit{A}}},\overline{\textbf{\textit{B}}})$, as shown in Eq.\ref{eq3}.
\begin{align}
&h'(t) = \textit{\textbf{A}}h(t) + \textbf{\textit{B}}x(t)\quad (\mathrm{1a}), \qquad y(t) = \textbf{\textit{C}}h(t)\quad (\mathrm{1b}), \label{eq1}\\   
&\overline{\textit{\textbf{A}}} = \exp(\Delta \textbf{\textit{A}}), \quad \overline{\textbf{\textit{B}}} = (\Delta \textbf{\textit{A}})^{-1}(\exp(\Delta \textbf{\textit{A}}) - \textit{\textbf{I}}) \cdot \Delta \textbf{\textit{B}},\label{eq2} \\   
&h_t = \overline{\textbf{\textit{A}}}h_{t-1} + \overline{\textbf{\textit{B}}}x_t\quad (\mathrm{3a}), \qquad y_t = \textit{\textbf{C}}h_t\quad (\mathrm{3b}). \label{eq3} 
\end{align}
where $\textbf{\textit{A}}\in \mathbb{R}^{N\times N}$, $\textbf{\textit{B}}\in \mathbb{R}^{N\times 1}$, $\textbf{\textit{C}}\in \mathbb{R}^{1\times N}$, and $\Delta$ is a time-scale parameter.
Mamba~\cite{gu2023mamba}, one of the most recent selective SSMs, introduces a gated selective mechanism to propagate or eliminate selected information based on the current state, significantly improving the content-reasoning performance. Specifically, Mamba changes the model from time-invariant to time-varying via converting parameters $\Delta$, \textbf{\textit{B}}, \textbf{\textit{C}} into input-dependent functions.

\begin{figure*}
\begin{center}
\includegraphics[width=13cm]{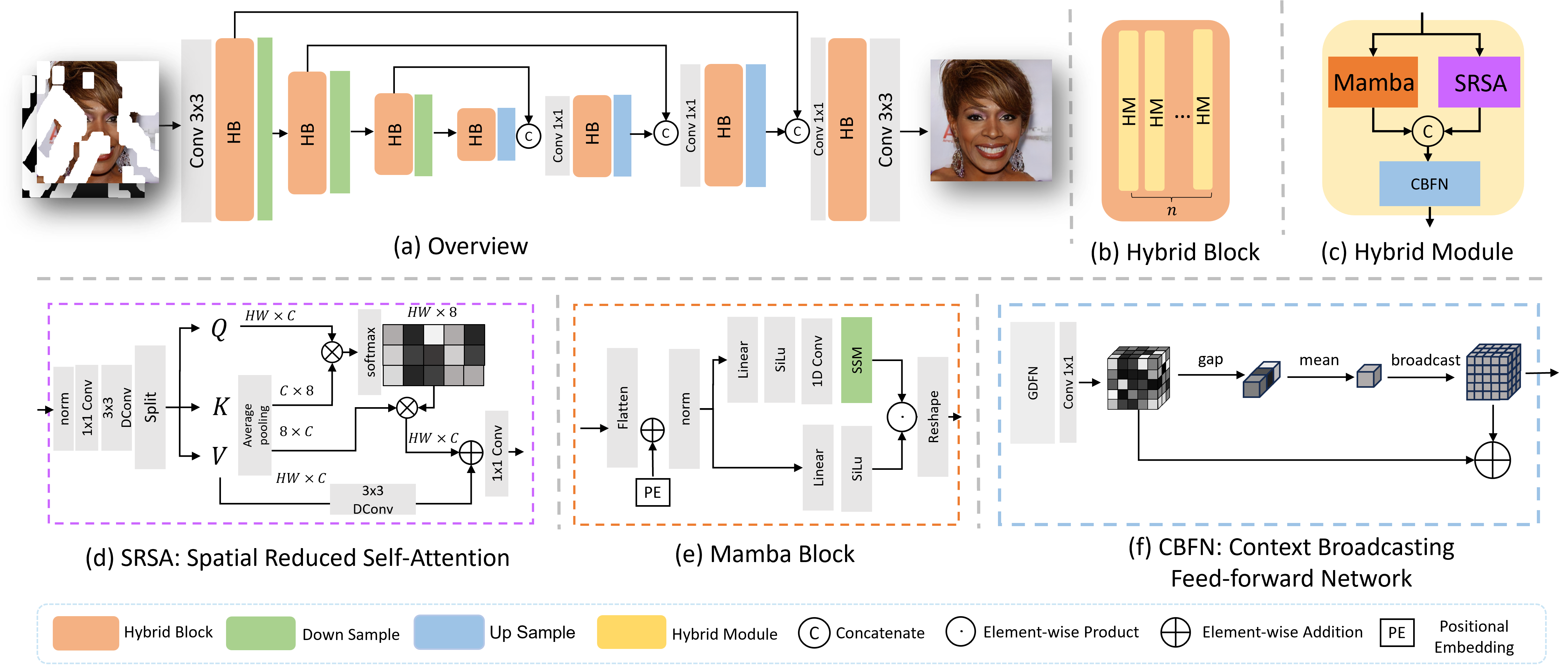}
\end{center}
   \caption{(a) The architecture overview of the proposed \name{}. (b) The Hybrid Block is composed of $n$ proposed Hybrid Modules. (c) The proposed Hybrid Module, consisted of a Mamba Block, a Spatial Reduced Self-Attention and a Context Broadcasting Feed-forward Network. (d) The Spatial Reduced Self-Attention provides spatial awareness. (e) The Mamba Block captures pixel-level interaction. (f) The Context Broadcasting Feed-forward Network transfers the features.}
\label{fig:overview}
\end{figure*}


\section{Method}
\label{sec:method}
The overall pipeline of the proposed \name{} is illustrated in Fig.~\ref{fig:overview}, which is a U-Net shape architecture formed with 7 Hybrid Blocks. Formally, the masked image ${I}_{masked}\in \mathbb{R}^{H\times W\times 3}$ concatenated with a mask ${M}\in \mathbb{R}^{H\times W\times 1}$ as the input ${I}_{in}$. We first use an overlapped convolution to embed ${I}_{in}$, then feed into the following 7 HBs with 3 times downsampling and 3 times upsampling. At the end, one convolution layer projects the final output ${I}_{out}$. Each Hybrid Block consists of $n$ Hybrid Modules (HM), as shown in Fig.~\ref{fig:overview} (b), where $n$ is the number of HMs. Each HM has a Transformer block, a Mamba block and a Context Broadcasting Feedforward Network (CBFN), which will be detailed in section~\ref{sec:method-hm}.

\subsection{Hybrid Module}
\label{sec:method-hm}
Each of the seven Hybrid Modules involves a pair of SRSA (Spatial Reduced Self-Attention) and Mamba modules for capturing long-range dependency, followed by a Context Broadcasting Feedforward Network (CBFN) to enhance the local context and control data flow consistency.
\\
\\
\noindent \textbf{Spatial Reduced Self-Attention.}
We introduce the Spatial Reduced Self-Attention (SRSA) module, designed to leverage the capability of the transformer for capturing global correlation while enriching local context detail.  
Specifically, given a input feature $F$, we first apply layer normalization followed by a $1 \times 1$ convolution and a $3\times3$ depth-wise convolution to extract the local features:
\begin{equation}
\begin{gathered}
F'=DConv_{3\times 3} (Conv_{1\times 1}(LayerNorm(F))).
\end{gathered}
\label{equ:srsa1}
\end{equation}

The feature $F'$ is then split along the channel dimensions to form the Query $Q$, Key $K$ and Value $V$. To address the traditional quadratic computational complexity of self-attention, we share the idea with PVTv2~\cite{wang2022pvt} to adopt average pooling for $K$ and $V$ to a fixed dimension.
\begin{equation}
\begin{gathered}
K', V' = AvgPool(K), AvgPool(V),\\
Att = softmax(K'\cdot Q),
\end{gathered}
\label{equ:srsa2}
\end{equation}
where $Att$ is the attention map. In this work, the spatial dimension is reduced to $8$. After multiplying $Att$ and $V'$, we get the initial output $F''$. To further enhance local context, we incorporate a Local Enhancement operation $LE(V)$ as proposed in \cite{ren2022shunted}, which is implemented using a $3\times 3$ depth-wise convolution, to effectively balances capturing extensive global interactions with detailed local features. After an element-wise addition, the final output of SRSA is:
\begin{equation}
\begin{gathered}
LE(V) = DConv_{3\times 3}(V),\\
Output_{srsa} = LE(V) + Att \cdot V',
\end{gathered}
\label{equ:srsa3}
\end{equation}
\\
\\
\noindent \textbf{Mamba with Positional Embedding.}
Mamba showcases a strong capacity to handle long sequence data with linear computational complexity, making it highly effective for modeling interactions between adjacent pixels. In this work, we propose leveraging the Mamba module to modeling the flattened feature, thereby capturing long-range dependency at the pixel level, which is expensive to capture by self-attention. To adapt Mamba more aptly for vision tasks and enhance its ability to maintain positional awareness, we incorporate positional embedding into the module.

Within the Mamba module, given an input feature $F$ with the shape of $(B,C,H,W)$, the process begins by flatting and transposing it to $(B,C,L)$, where $L=H\times W$:

\begin{equation}
\begin{gathered}
F' = transpose(reshape(F, (B,C,L))).
\end{gathered}
\label{equ:mamba1}
\end{equation}
Subsequently, we introduce cosine positional embedding~\cite{vaswani2017attention} to the transformed feature, enhancing the capacity to maintain positional awareness:
\begin{equation}
\begin{gathered}
F'' = F'+ PE(L).
\end{gathered}
\label{equ:mamba2}
\end{equation}
After applying layer normalization, mamba implements a gated mechanism to further refine the feature representation. The body branch involves a linear layer, a SiLU activation function~\cite{hendrycks2016gaussian}, $1D$ convolutional layer and the SSM (State Space Sequence Models) layer. 
\begin{equation}
\begin{gathered}
F_{body} = SSM(Conv1D(SiLU(Linear(F''))))
\end{gathered}
\label{equ:mamba3}
\end{equation}
The gate branch involves a linear layer and a SiLU activation function~\cite{hendrycks2016gaussian}. After the gate branch re-weight the body branch, the output will be reshaped to the shape of $(B,C,H,W)$:

\begin{equation}
\begin{gathered}
G = SiLU(Linear(F'')),\\
F_{gated} = G\cdot F_{body},\\
Output_{mamba} = reshape(F_{gated}, (B,C, H, W)),
\end{gathered}
\label{equ:mamba2}
\end{equation}
where G is the gate matric, $F_{gated}$ is the output from gate mechanism, $Output_{mamba}$ is the final output from Mamba module.
\\
\\
\noindent \textbf{Context Broadcasting Feed-forward Network.}
We propose Context Broadcasting Feedforward Network (CBFN) by improving the Gated-Dconv Feed-Forward Network (GDFN)~\cite{zamir2022restormer}. The GDFN is recognized for its efficacy in enhancing local context through a gated mechanism with depth-wise convolution. To build upon this, our CBFN integrates a global processing stage post-GDFN. Specifically, we implement global average pooling followed by channel-wise averaging to obtain the overall mean value of the input feature $F$, denoted as $\mu = GlobalAvgPool(F)$, where $F$ is the output from GDFN. This $\mu$ is then broadcast to the dimensions of $F$ and added to it. The output of CBFN is represented as $F'$:
\begin{equation}
\begin{gathered}
F'=F + broadcast(\mu).
\end{gathered}
\label{equ:avg}
\end{equation}
This global processing is designed to facilitate the learning of dense interactions within the self-attention layers~\cite{hyeon2023scratching}, thereby enhancing the effectiveness of the Hybrid Module.

\subsection{Loss Functions}
To achieve superior inpainting outcomes, 
we adopt a multi-component loss strategy as delineated in the previous research~\cite{chen2024hint,nazeri2019edgeconnect,li2022misf}. This strategy includes an $\mathcal{L}_1$ loss, a style loss $\mathcal{L}_{\text {style}}$, a perceptual loss $\mathcal{L}_{\text {perc}}$ and an  adversarial loss $\mathcal{L}_{\text {adv}}$.
The composite loss function is formulated as:
\begin{equation}
\mathcal{L}_{all}({I}_{out}, {I_{gt}})= \alpha_1 \mathcal{L}_1 + \alpha_2 \mathcal{L}_{\text{style}} + \alpha_3 \mathcal{L}_{\text {perc}} + \alpha_4 \mathcal{L}_{\text {adv}},
\end{equation}
where ${I}_{out}$ and ${I}_{gt}$ are the reconstructed image and ground truth, respectively. $\alpha_1$=1, $\alpha_2$=250, $\alpha_3$=0.1, and $\alpha_4$=0.001 are the weighting factors for each component.

\section{Experiment Results}
\noindent \textbf{Datasets.}
We evaluate our \name{} on two diverse datasets, CelebA-HQ~\cite{karras2017progressive} and Places2-standard~\cite{zhou2017places}, to ensure a comprehensive comparison. CelebA-HQ is a dataset consisting of high-quality human face images. For CelebA-HQ, we train our model on the first 28000 images and reserve the remaining 2000 for testing. Places2 comprises a wide range of natural and indoor scene images. For Places2, we employ the standard training set, which includes 1.8 million images, and test on its validation set of 30000 images. We follow \cite{guo2021image, li2022misf, chen2024hint} to conduct all experiments with the widely used irregular mask~\cite{liu2018image} in three mask ratios.
\\
\\
\noindent \textbf{Implementation Details.}
All experiments are carried out on one Nvidia A100 GPU. During training, we adopt the Adam optimiser~\cite{kingma2020method} with ${\beta}_1$ = 0.9, ${\beta}_2$ = 0.999. The learning rate is set to $1\times 10^{-4}$ and the batch size is 4. In the Hybrid Blocks, the numbers of Hybrid Modules are [4,6,6,8,6,6,4].
\\
\\
\noindent \textbf{Evaluation Metrics.} We followed~\cite{li2022misf,chen2024hint} to evaluate the image generation quality with Peak Signal-to-Noise ratio (PSNR), Structural similarity (SSIM), L1, Frechet inception distance (FID) and Perceptual Similarity (LPIPS). We use these metrics as they offer insights into the quality of the images generated by the model. PSNR, SSIM, and L1 metrics evaluate reconstruction quality at the pixel level, assessing fine-grained details and structural context. FID quantifies the distributional differences between generated images and the original dataset. LPIPS is employed to reflect differences in human perception.

\newcommand{\PSNRT}{PSNR}
\newcommand{\SSIMT}{SSIM}
\newcommand{\LT}{L1}
\newcommand{\LPIPST}{LPIPS}
\newcommand{\FID}{FID}
\newcommand{\LowExposure}{0.01\%-20\%}
\newcommand{\MiddleExposure}{20\%-40\%}
\newcommand{\HighExposure}{40\%-60\%}
\newcommand{\AllExposure}{Default}

\newcommand{\Frst}[1]{{\textbf{#1}}}
\newcommand{\Scnd}[1]{{\underline{#1}}}

\begin{table*}[h]
\centering
\resizebox{\textwidth}{!}{
    \begin{tabular}{@{\extracolsep{3pt}}c c c c c c| c c c c c| c c c c c@{}}
    \hline\hline
    \textbf{CelebA-HQ}
    & \multicolumn{5}{c}{\LowExposure} & \multicolumn{5}{c}{\MiddleExposure} & \multicolumn{5}{c}{\HighExposure} \T\\
    Method &\PSNRT$\uparrow$ & \SSIMT$\uparrow$ & \LT$\downarrow$ & \FID$\downarrow$ & \LPIPST$\downarrow$ & \PSNRT$\uparrow$ & \SSIMT$\uparrow$ & \LT$\downarrow$ & \FID$\downarrow$ & \LPIPST$\downarrow$ & \PSNRT$\uparrow$ & \SSIMT$\uparrow$ & \LT$\downarrow$ & \FID$\downarrow$ & \LPIPST$\downarrow$  \\
    \hline
    DeepFill v1 \cite{yu2018generative} & 34.2507 & 0.9047 & 1.7433 & 2.2141 & 0.1184 & 26.8796 & 0.8271 & 2.3117 & 9.4047 & 0.1329 & 21.4721 & 0.7492 & 4.6285 & 15.4731 & 0.2521\T\\

    DeepFill v2 \cite{yu2019free} & 34.4735 & 0.9533 & 0.5211 & 1.4374 & 0.0429 & 27.3298 & 0.8657 & 1.7687 & 5.5498 & 0.1064 & 22.6937 & 0.7962 & 3.2721 & 8.8673 & 0.1739\\

    WaveFill \cite{yu2021wavefill} & 31.4695 & 0.9290 & 1.3228 & 6.0638 & 0.0802 & 27.1073 & 0.8668 & 2.1159 & 8.3804 & 0.1231 & 23.3569 & 0.7817 & 3.5617 & 13.0849 & 0.1917\\
    RePaint~\cite{Lugmayr_2022_CVPR} & -&-& -& -& -& -& -& -& -& -& 21.8321 & 0.7791 & 3.9427 & 8.9637 & 0.1943 \\
    LaMa \cite{suvorov2022resolution} & {35.5656} & 0.9685 & 0.4029 & 1.4309 & 0.0319 & {28.0348} & 0.8983 & {1.3722} & 4.4295 & 0.0903 & \Scnd{23.9419} & {0.8003} & {2.8646} & 8.4538 & 0.1620\\


    MISF \cite{li2022misf} & 35.3591 & 0.9647 & 0.4957 & 1.2759 & 0.0287 & 27.4529 & 0.8899 & 2.0118 & 4.7299 & 0.1176 & 23.4476 & 0.7970 & 3.4167 & 8.1877 & 0.1868\\
    
    MAT \cite{li2022mat}  & 35.5466 & {0.9689} & {0.3961} & {1.2428} & {0.0268} & 27.6684 & {0.8957} & 1.3852 & {3.4677} & {0.0832} & 23.3371 & 0.7964 & 2.9816 & {5.7284} & {0.1575}\\

    CMT \cite{ko2023continuously} & \Scnd{36.0336} & \Scnd{0.9749} & \Scnd{0.3739} & \Scnd{1.1171} & \Scnd{0.0261} & \Scnd{28.1589} & \Scnd{0.9109} & \Scnd{1.2938} & \Scnd{3.3915} & \Scnd{0.0817} & 23.8183 & \Scnd{0.8141} & \Scnd{2.8025} & \Scnd{5.6382} & \Scnd{0.1567}\B\\
    \hline
    Ours & \Frst{36.7394} & \Frst{0.9737} & \Frst{0.3614} & \Frst{1.1142} & \Frst{0.0229} & \Frst{28.8098} & \Frst{0.9112} & \Frst{1.2413} &\Frst{3.3890} & \Frst{0.0722} & \Frst{24.3784} & \Frst{0.8220} & \Frst{2.6739} & \Frst{5.6041} & \Frst{0.1402} \B\T\\
    \hline\hline


    \textbf{Places2} & \multicolumn{5}{c}{\LowExposure} & \multicolumn{5}{c}{\MiddleExposure} & \multicolumn{5}{c}{\HighExposure} \T\\
     Method & \PSNRT$\uparrow$ & \SSIMT$\uparrow$ & \LT$\downarrow$ &\FID$\downarrow$ & \LPIPST$\downarrow$ & \PSNRT$\uparrow$ & \SSIMT$\uparrow$ & \LT$\downarrow$ &\FID$\downarrow$ & \LPIPST$\downarrow$ & \PSNRT$\uparrow$ & \SSIMT$\uparrow$ & \LT$\downarrow$ &\FID$\downarrow$ & \LPIPST$\downarrow$ \\
    \hline

    DeepFill v1\cite{yu2018generative} & 30.2958 & 0.9532 & 0.6953 & 26.3275 & 0.0497 & 24.2983 & 0.8426 & 2.4927 & 31.4296 & 0.1472 & 19.3751 & 0.6473 & 5.2092 & 46.4936 & 0.3145\T\\

    DeepFill v2\cite{yu2019free} & 31.4725 & 0.9558 & 0.6632 & 23.6854 & 0.0446 & 24.7247 & 0.8572 & 2.2453 & 27.3259 & 0.1362 & 19.7563 & 0.6742 & 4.9284 & 36.5458 & 0.2891\\

    CTSDG \cite{guo2021image}  & 32.1110 & 0.9565 & 0.6216 & 24.9852 & 0.0458 & 24.6502 & 0.8536 & 2.1210 &  29.2158 & 0.1429 & 20.2962 & 0.7012 & 4.6870 & 37.4251 & 0.2712\\

    WaveFill \cite{yu2021wavefill} & 29.8598 & 0.9468 & 0.9008 &30.4259 & 0.0519 & 23.9875 & 0.8395 & 2.5329 & 39.8519 & 0.1365 & 18.4017 & 0.6130 & 7.1015 & 56.7527 & 0.3395\\
    LDM~\cite{Rombach_2022_CVPR} & -&-& -& -& -& -& -& -& -& -& 19.6476 & 0.7052 & 4.6895 & 27.3619 & 0.2675\\
    Stable Diffusion$^*$ & -&-& -& -& -& -& -& -& -& -& 19.4812 & 0.7185 & 4.5729 & 27.8830 & 0.2416\\
    WNet \cite{zhang2022w} & 32.3276 & 0.9372 & 0.5913 & 20.4925 & 0.0387 & 25.2198 & 0.8617 & 2.0765 & 24.7436 & 0.1136 & 20.4375 & 0.6727 & 4.6371 & 32.6729 & 0.2416\\

    MISF~ \cite{li2022misf}  & \Scnd{32.9873} & \Scnd{0.9615} & \textbf{0.5931} & 21.7526 & 0.0357 & \textbf{25.3843} & \Scnd{0.8681} & \Scnd{1.9460} & 30.5499  & 0.1183 & \textbf{20.7260} & 0.7187 & 4.4383 & 44.4778 & 0.2278 \\

    LaMa \cite{suvorov2022resolution}  & 32.4660 & 0.9584 & 0.5969 & \textbf{14.7288} & \Scnd{0.0354} & 25.0921 & 0.8635 & 2.0048 & \textbf{22.9381} & \textbf{0.1079} & 20.6796 & \Scnd{0.7245} & \Scnd{4.4060} & \textbf{25.9436} & \textbf{0.2124}\\   

    CMT \cite{ko2023continuously} & 32.5765 & 0.9624 & 0.5915 & 22.1841 & 0.0364 & 24.9765 & 0.8666 & 2.0277 & 32.0184 & 0.1184 &  20.4888 & 0.7111 &  4.5484& 35.1688 & 0.2378 \B\\

    \hline
    Ours & \textbf{32.9940} & \textbf{0.9672} & \Scnd{0.5950} & \Scnd{15.3980} & \textbf{0.0334} & \Scnd{25.3278} & \textbf{0.8756} & \textbf{1.9404} & \Scnd{23.7109} & \Scnd{0.1106} &  \Scnd{20.7022} & \textbf{0.7319} &  \textbf{4.3379} & \Scnd{26.9155} & \Scnd{0.2372} \T\B\\
    \hline\hline
    \multicolumn{16}{l}{\large $^*$: The officially released Stable Diffusion inpainting model pretrained on high-quality LAION-Aesthetics V2 5+ dataset.}\T\\ 
    \end{tabular}
}
\caption{Quantitative comparison against state-of-the-art methods on CelebA-HQ (top), and Places2 (bottom). The best and second-best are indicated by \textbf{Bold} and \underline{underline}, respectively.}
\label{tab:long}
\end{table*}

\subsection{Comparison with State of the Art}
\begin{figure*}[t] \centering
    \makebox[0.1\textwidth]{\tiny Input}
    \makebox[0.1\textwidth]{\tiny WaveFill~\cite{yu2021wavefill}}
    \makebox[0.1\textwidth]{\tiny LAMA~\cite{wan2021high}}
    \makebox[0.1\textwidth]{\tiny MISF~\cite{li2022misf}}
    \makebox[0.1\textwidth]{\tiny MAT~\cite{li2022mat}}
    \makebox[0.1\textwidth]{\tiny RePaint~\cite{Lugmayr_2022_CVPR}}
    \makebox[0.1\textwidth]{\tiny CMT~\cite{ko2023continuously}}
    \makebox[0.1\textwidth]{\tiny Ours}
    \makebox[0.1\textwidth]{\tiny GT}
    \\
    \includegraphics[width=0.1\textwidth]{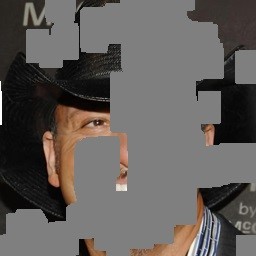}
    \includegraphics[width=0.1\textwidth]{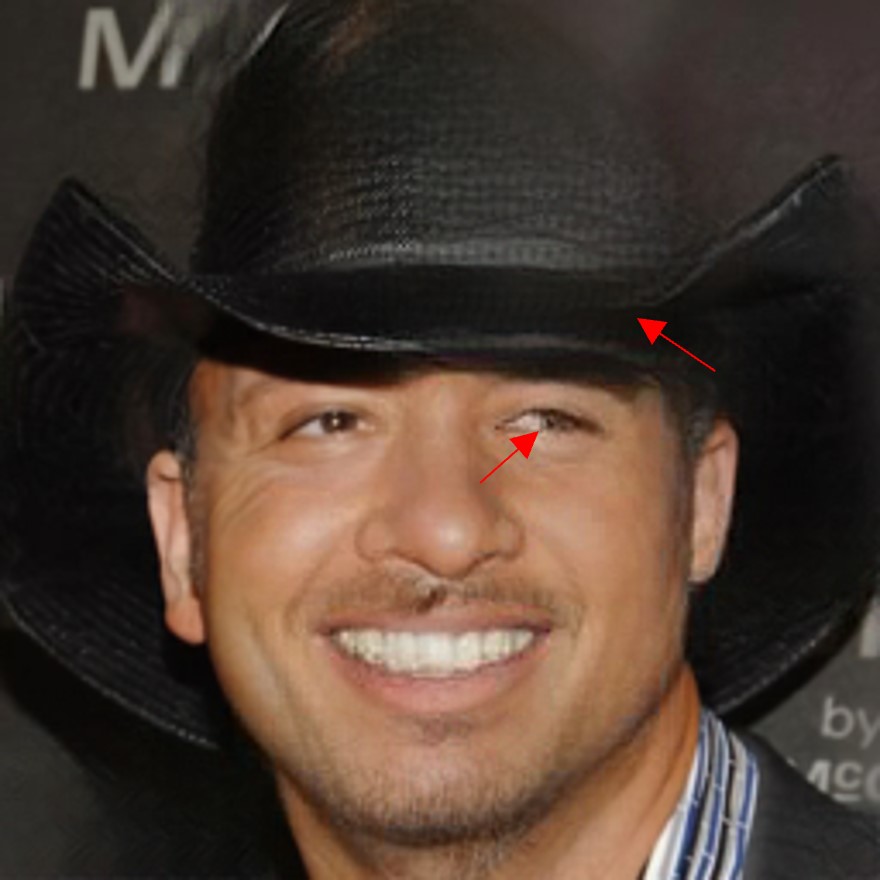}
    \includegraphics[width=0.1\textwidth]{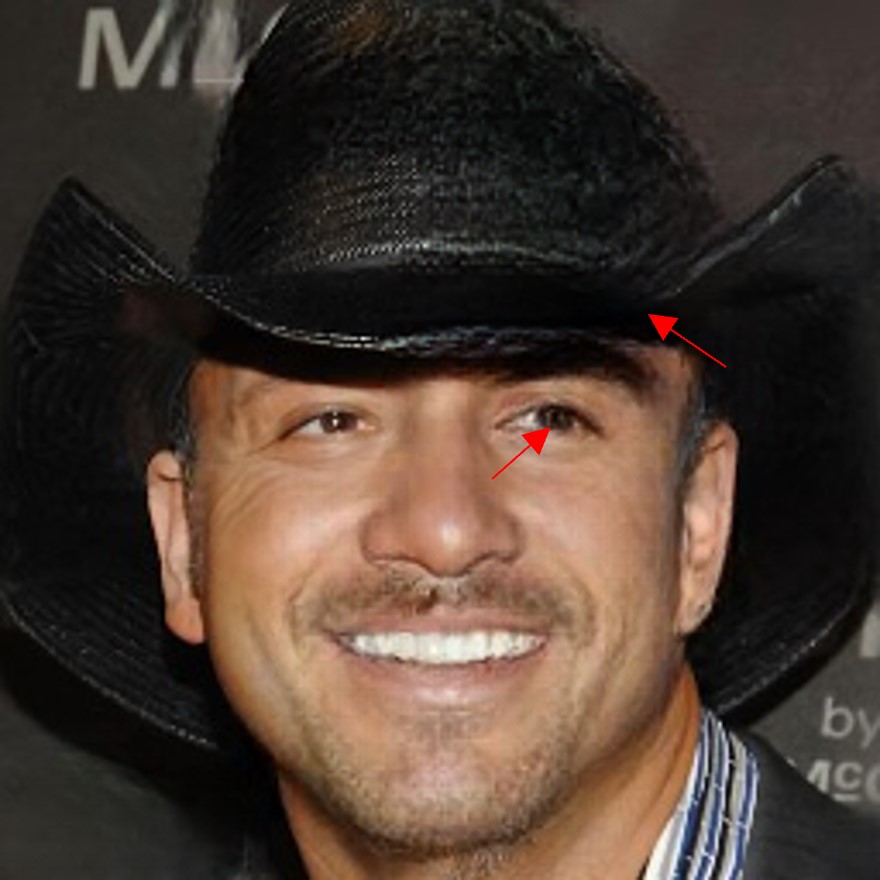}
    \includegraphics[width=0.1\textwidth]{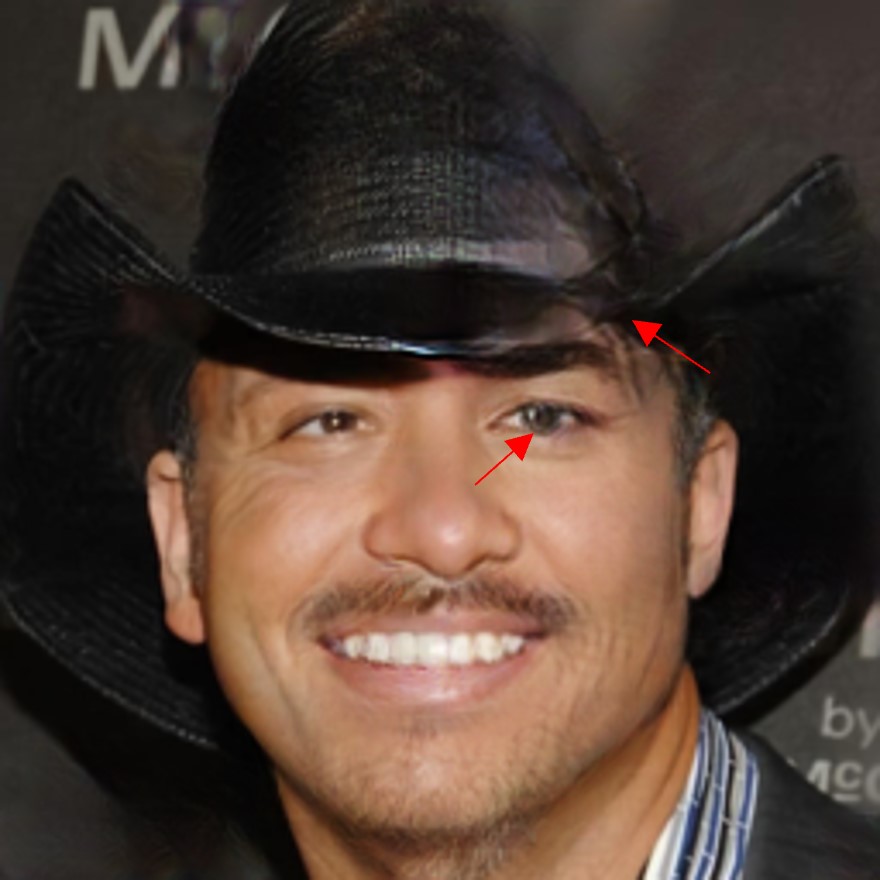}
    \includegraphics[width=0.1\textwidth]{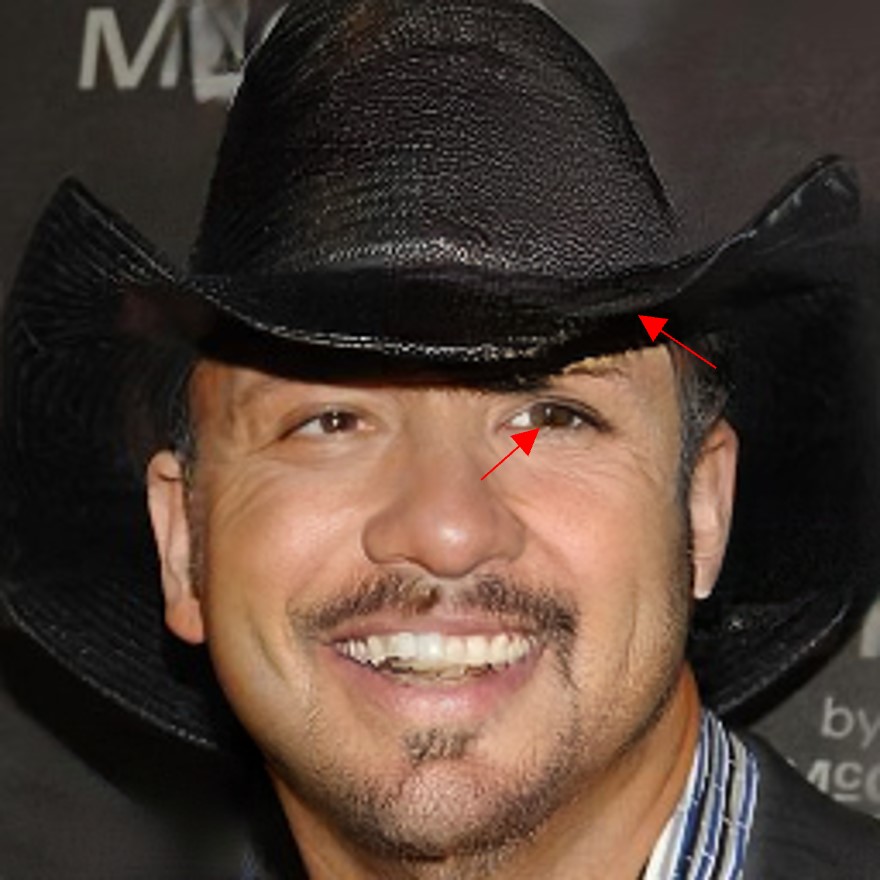}
    \includegraphics[width=0.1\textwidth]{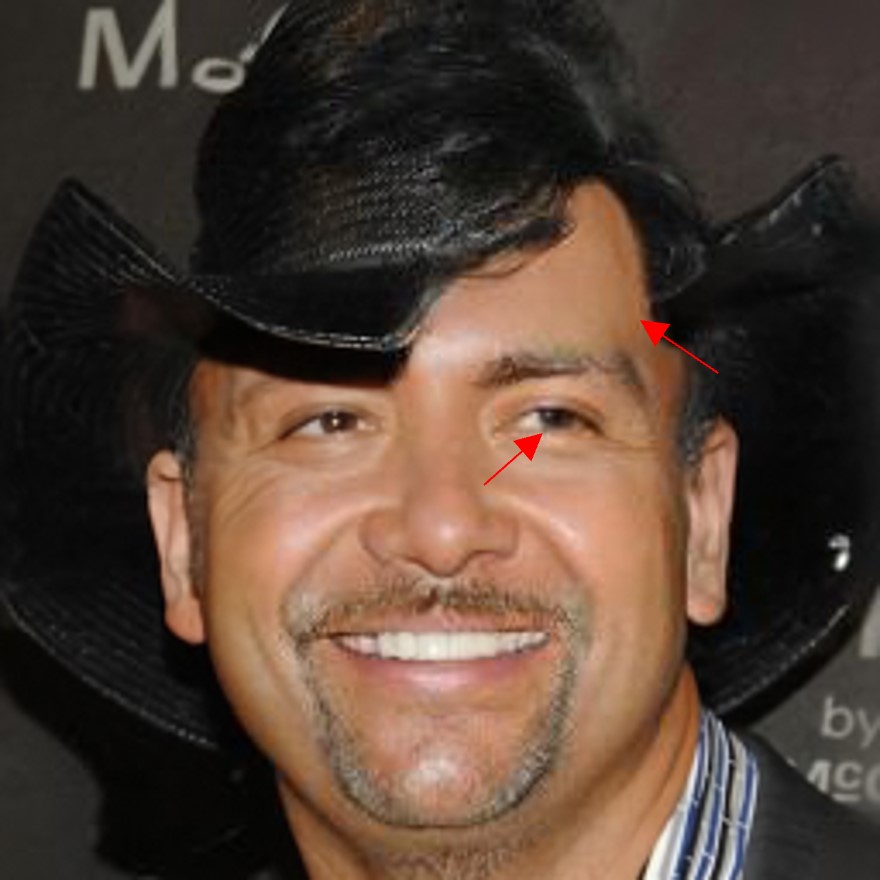}
    \includegraphics[width=0.1\textwidth]{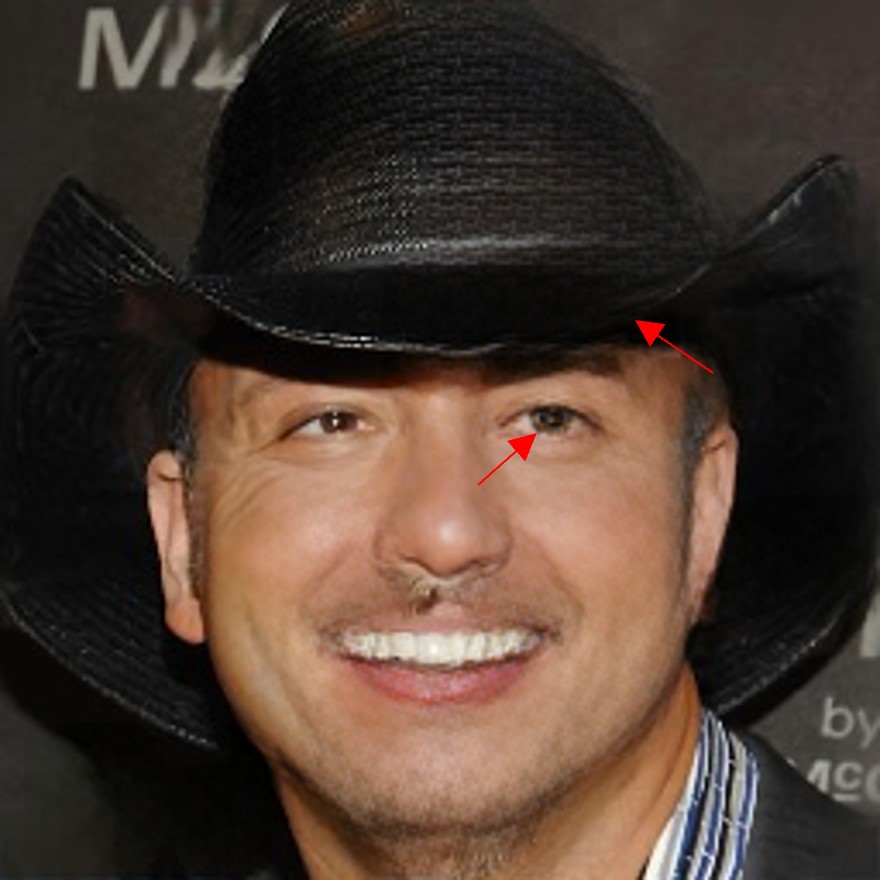}
    \includegraphics[width=0.1\textwidth]{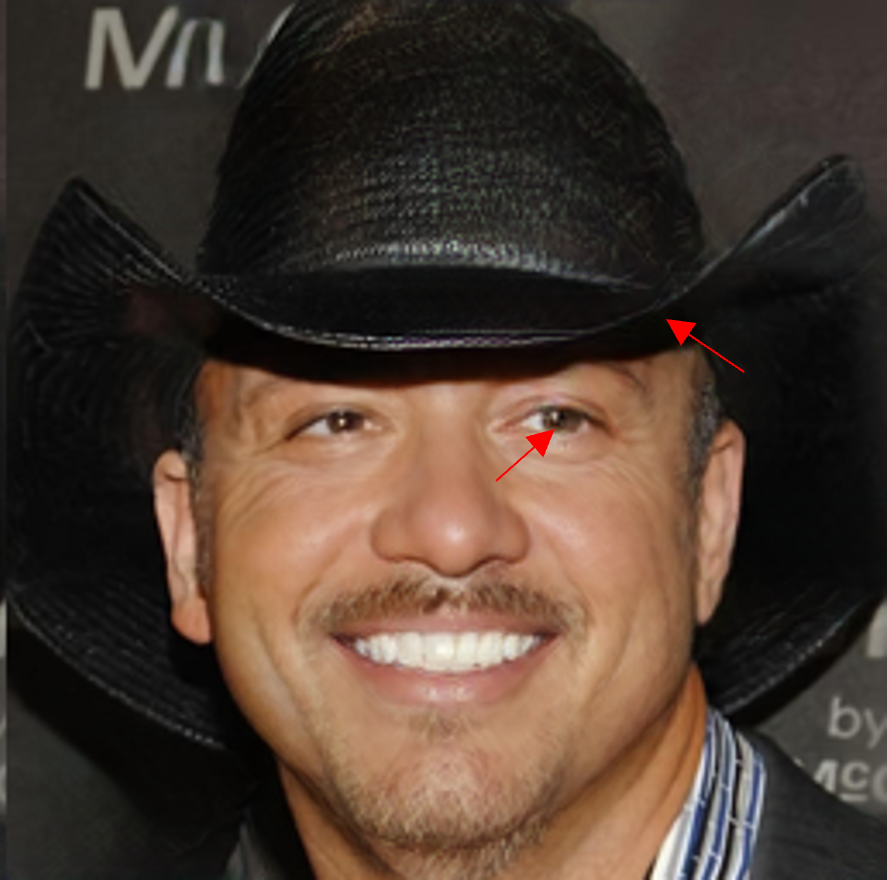}
    \includegraphics[width=0.1\textwidth]{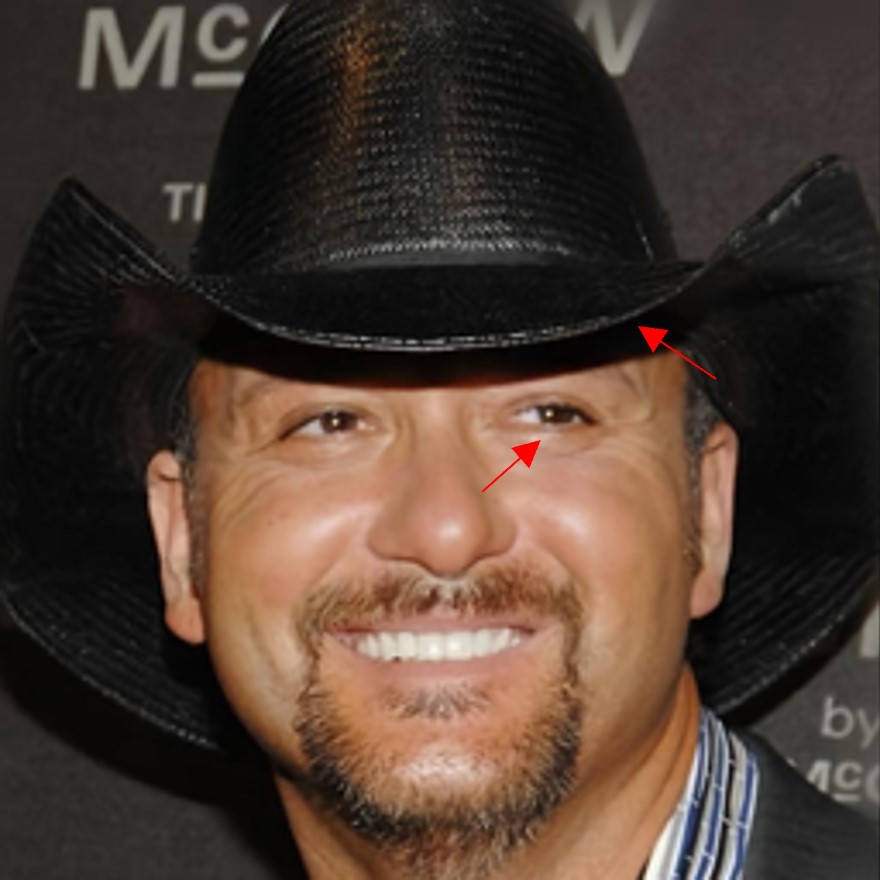}
    \\
    \includegraphics[width=0.1\textwidth]{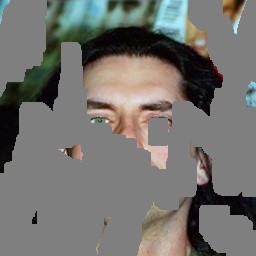}
    \includegraphics[width=0.1\textwidth]{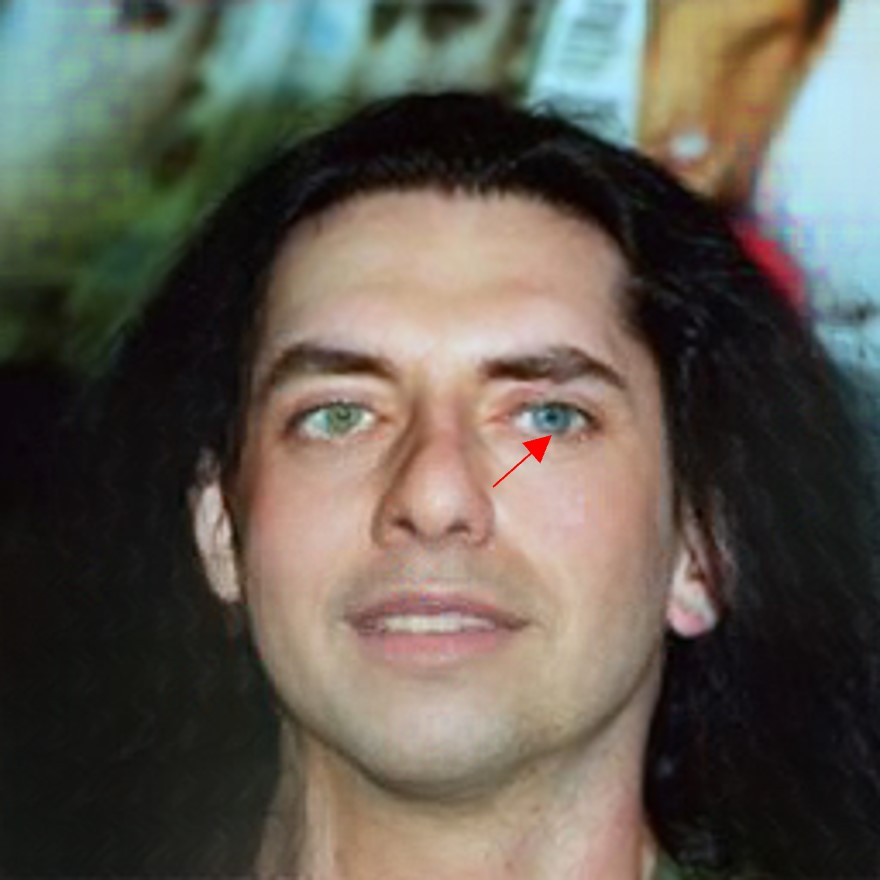}
    \includegraphics[width=0.1\textwidth]{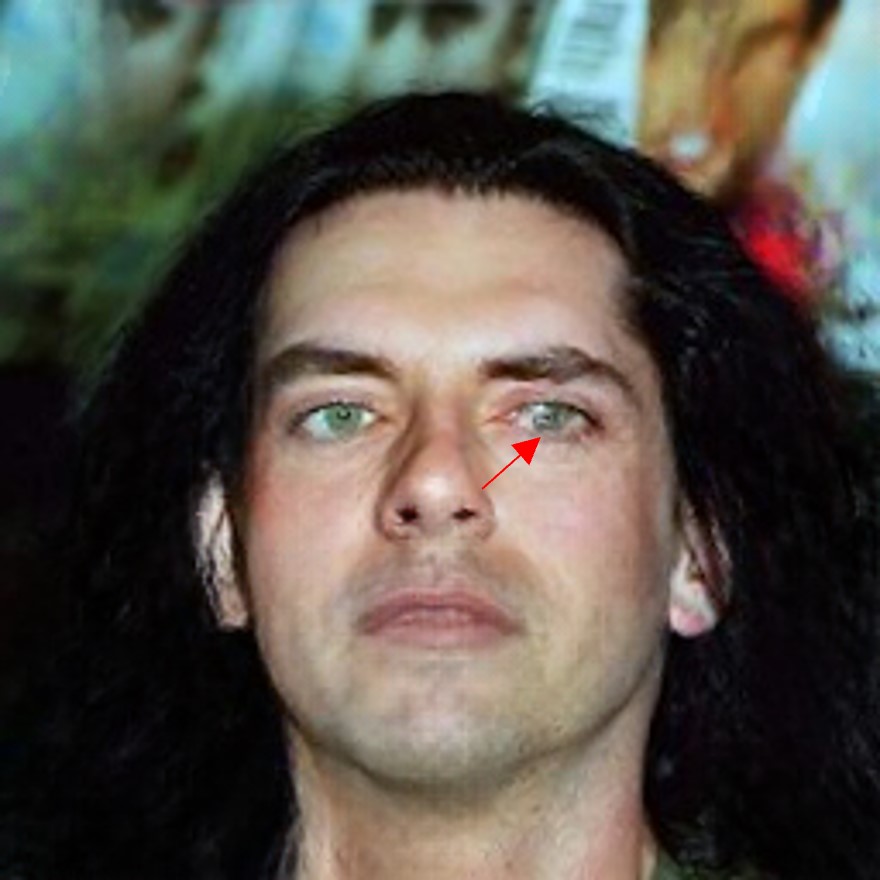}
    \includegraphics[width=0.1\textwidth]{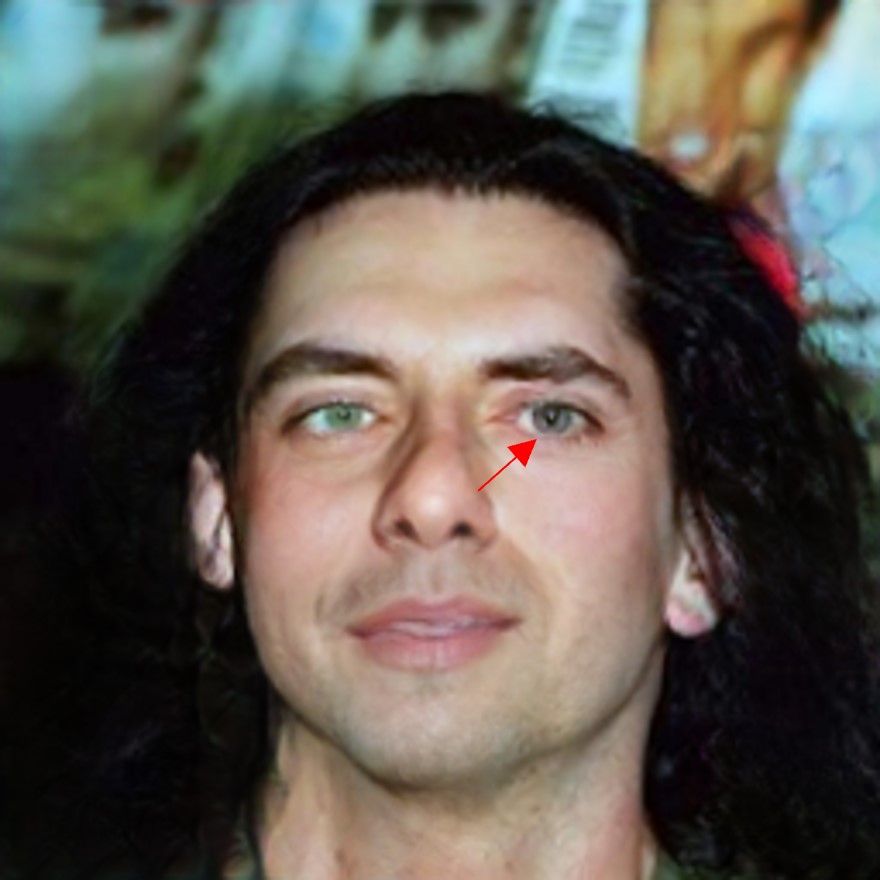}
    \includegraphics[width=0.1\textwidth]{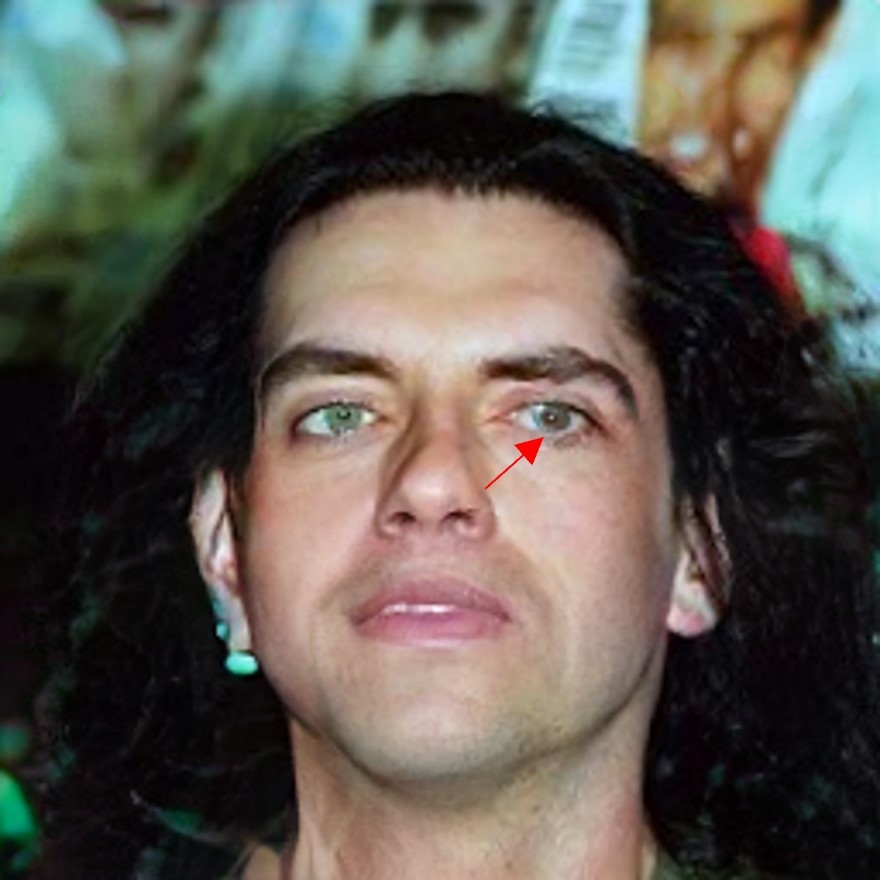}
    \includegraphics[width=0.1\textwidth]{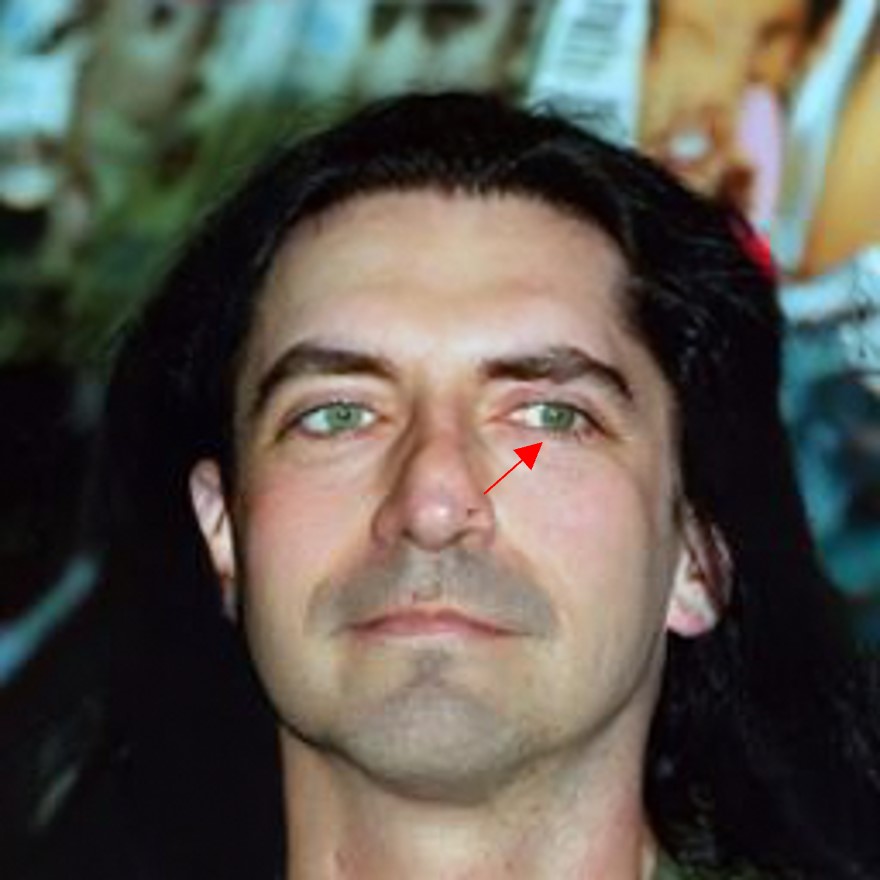}
    \includegraphics[width=0.1\textwidth]{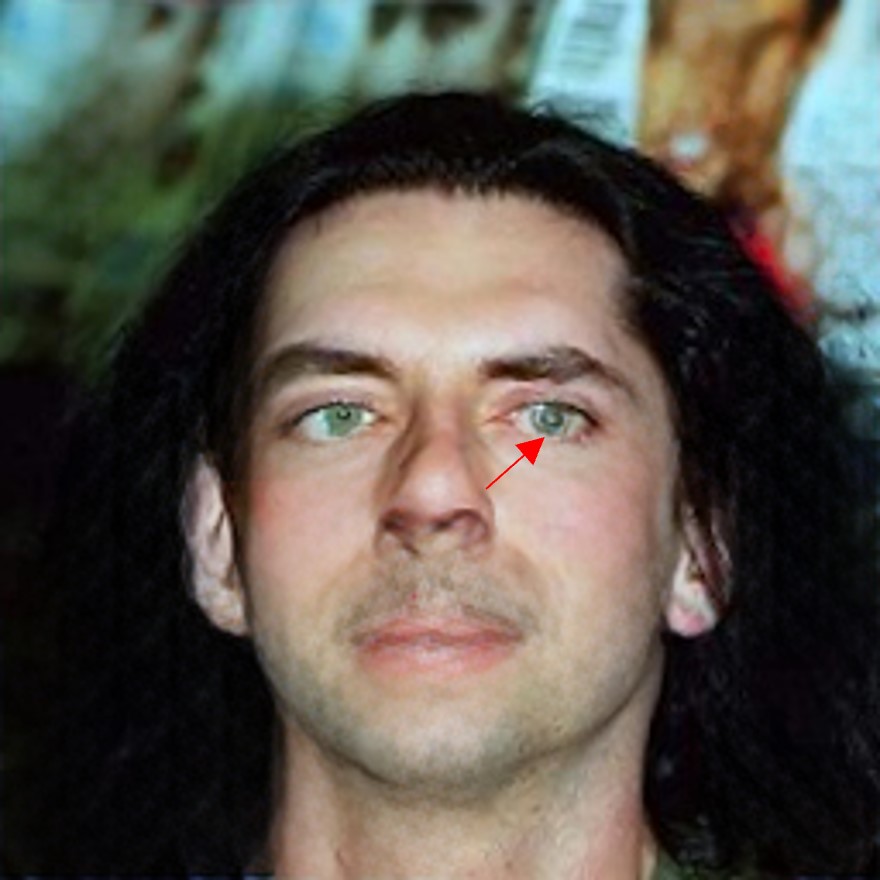}
    \includegraphics[width=0.1\textwidth]{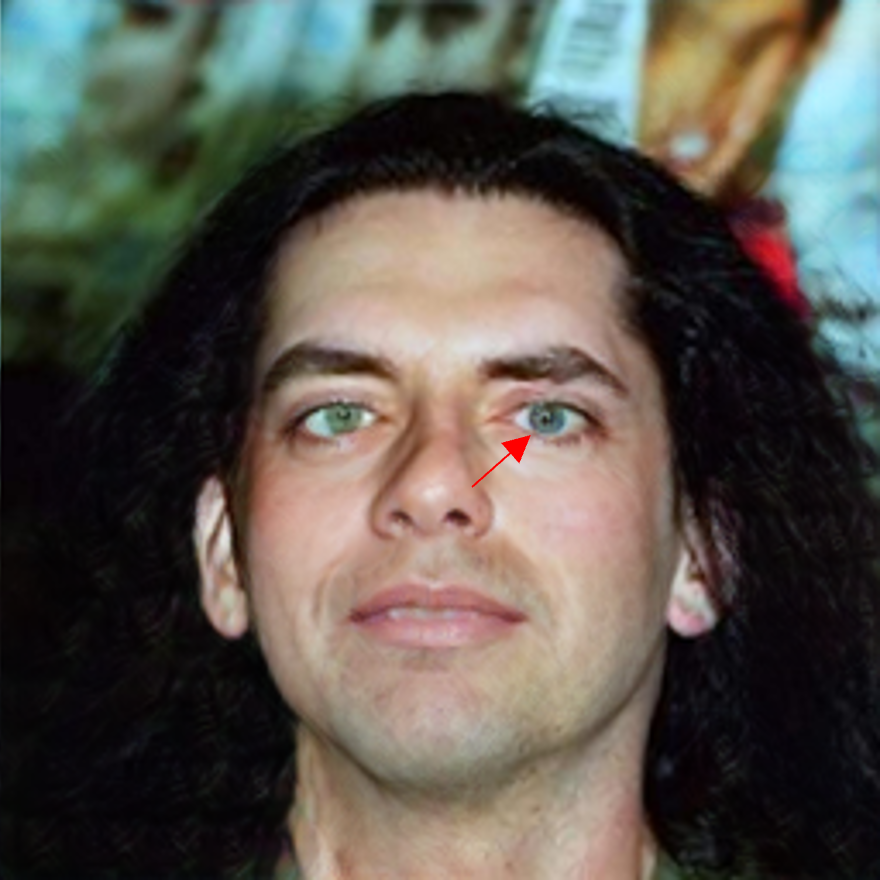}
    \includegraphics[width=0.1\textwidth]{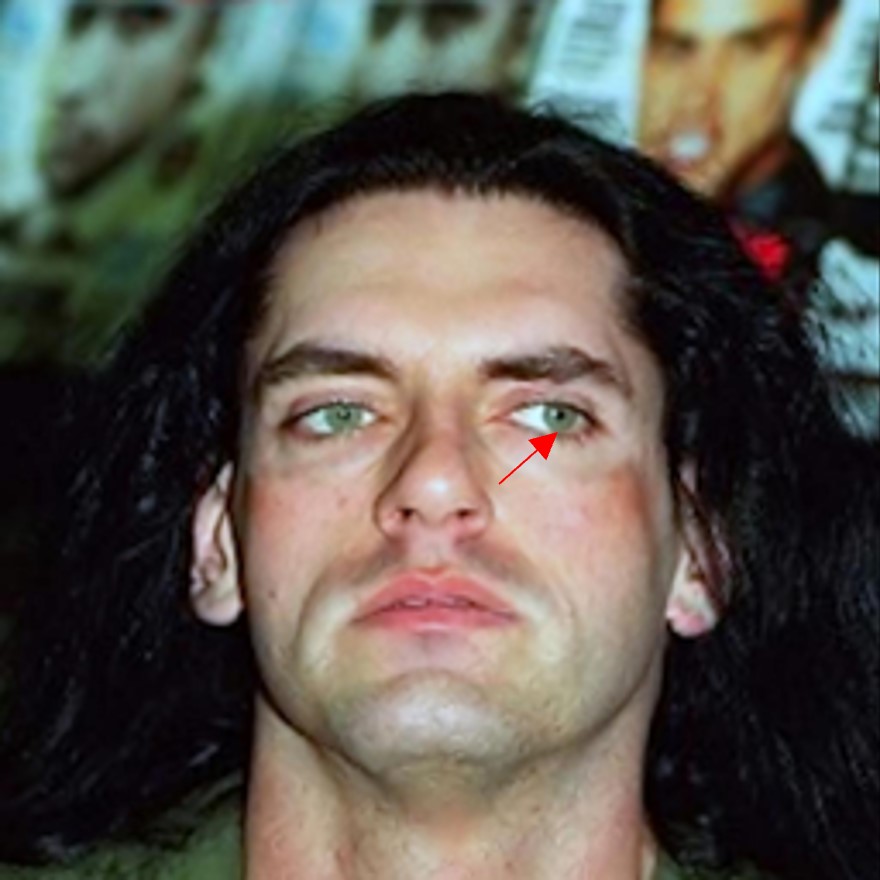}
    \makebox[0.1\textwidth]{\tiny Input}
    \makebox[0.1\textwidth]{\tiny WaveFill~\cite{yu2021wavefill}}
    \makebox[0.1\textwidth]{\tiny LAMA~\cite{wan2021high}}
    \makebox[0.1\textwidth]{\tiny MISF~\cite{li2022misf}}
    \makebox[0.1\textwidth]{\tiny MAT~\cite{li2022mat}}
    \makebox[0.1\textwidth]{\tiny RePaint~\cite{Lugmayr_2022_CVPR}}
    \makebox[0.1\textwidth]{\tiny CMT~\cite{ko2023continuously}}
    \makebox[0.1\textwidth]{\tiny Ours}
    \makebox[0.1\textwidth]{\tiny GT}
    \\
    \includegraphics[width=0.1\textwidth]{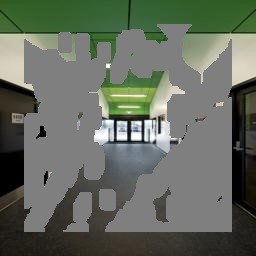}
    \includegraphics[width=0.1\textwidth]{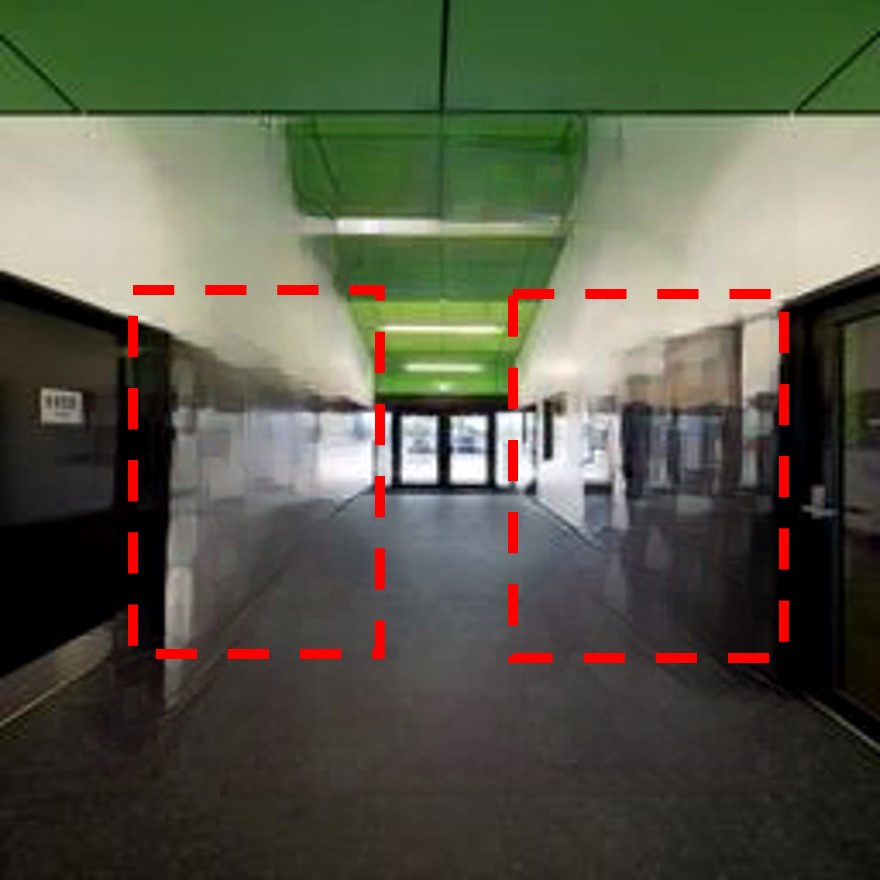}
    \includegraphics[width=0.1\textwidth]{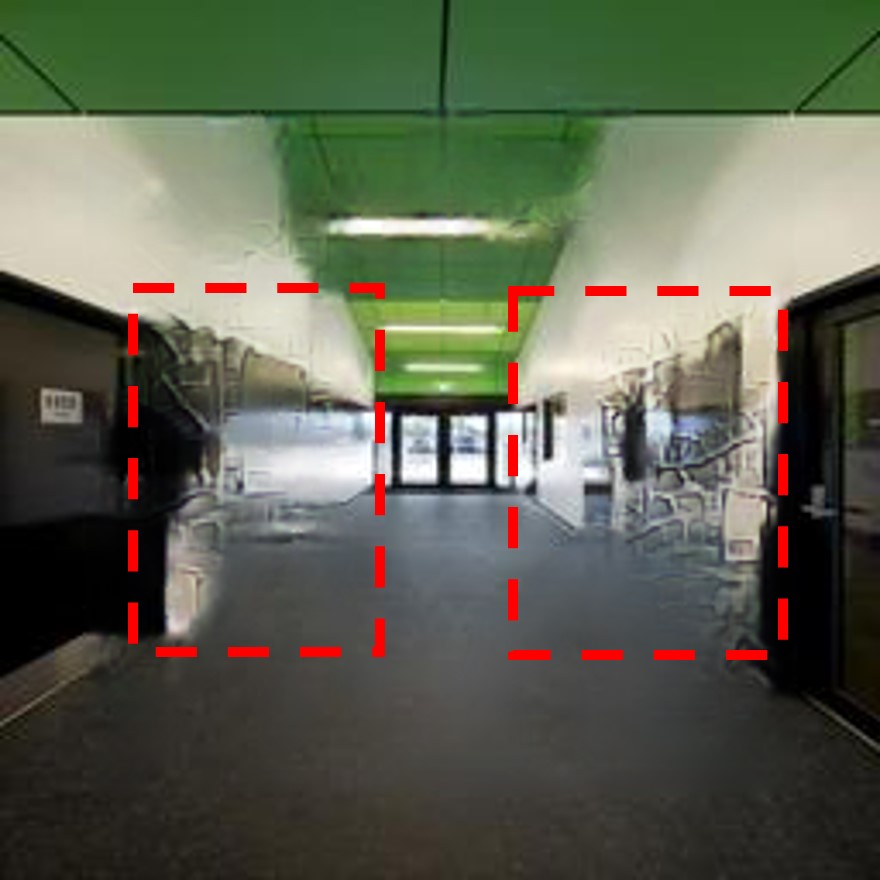}
    \includegraphics[width=0.1\textwidth]{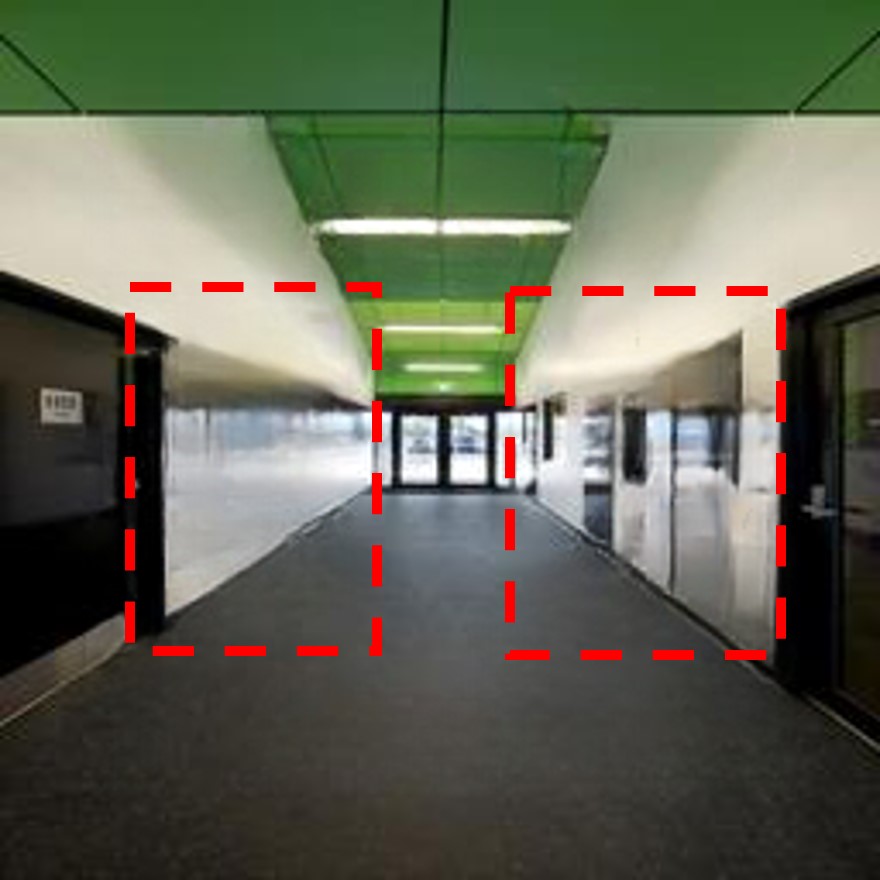}
    \includegraphics[width=0.1\textwidth]{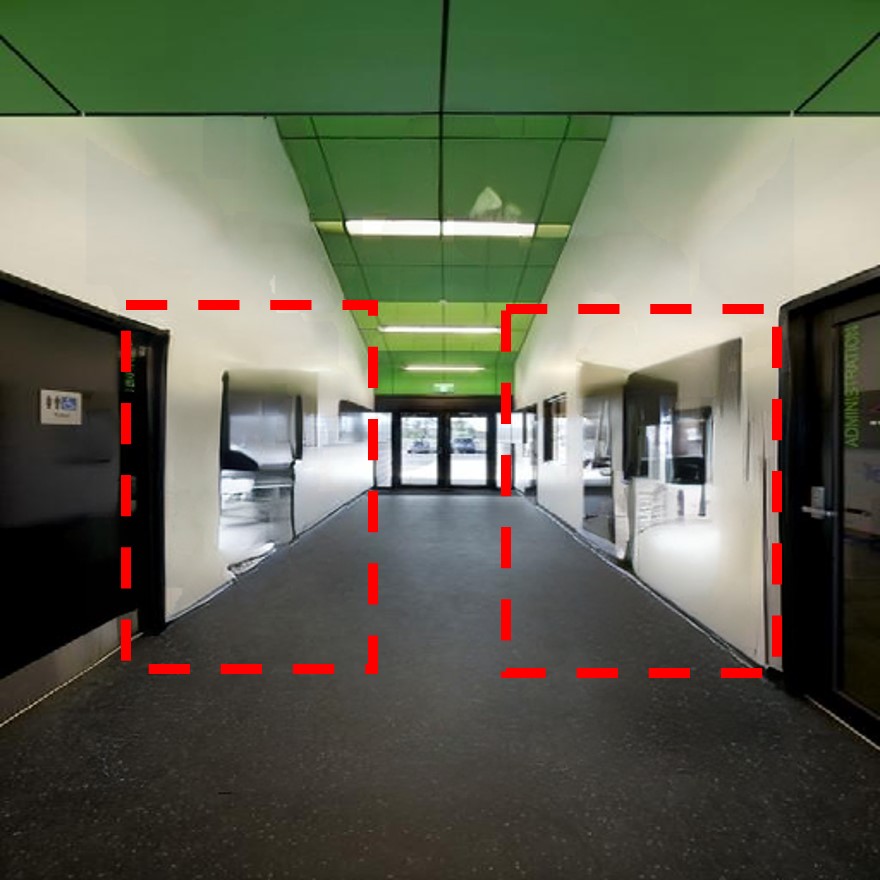}
    \includegraphics[width=0.1\textwidth]{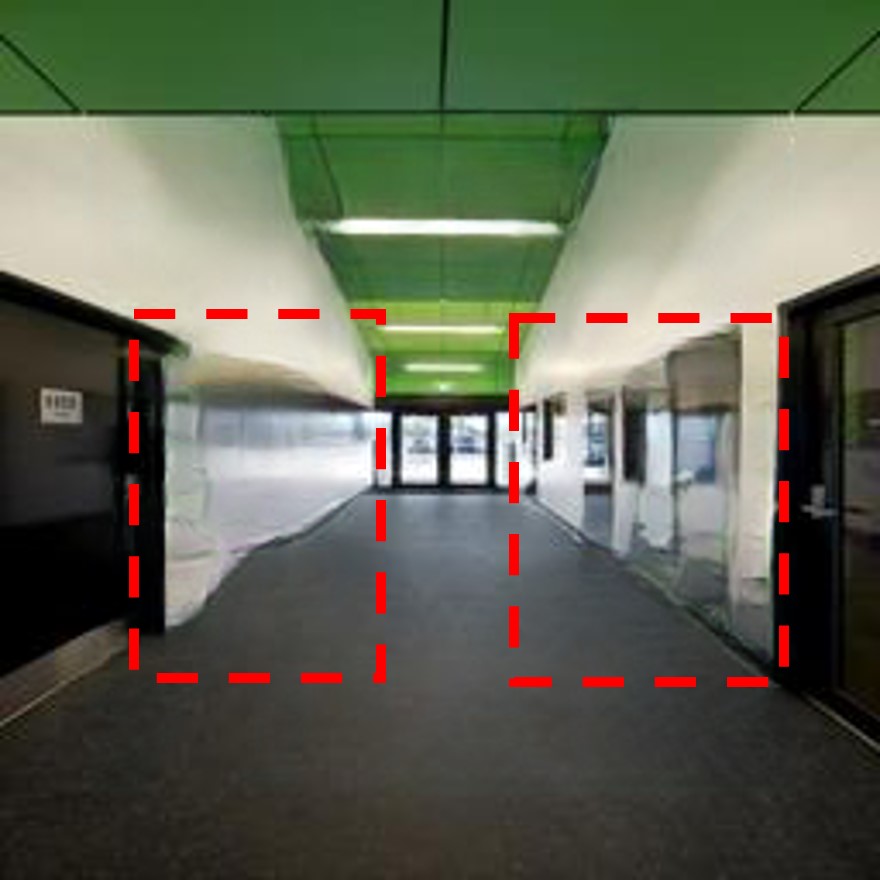}
    \includegraphics[width=0.1\textwidth]{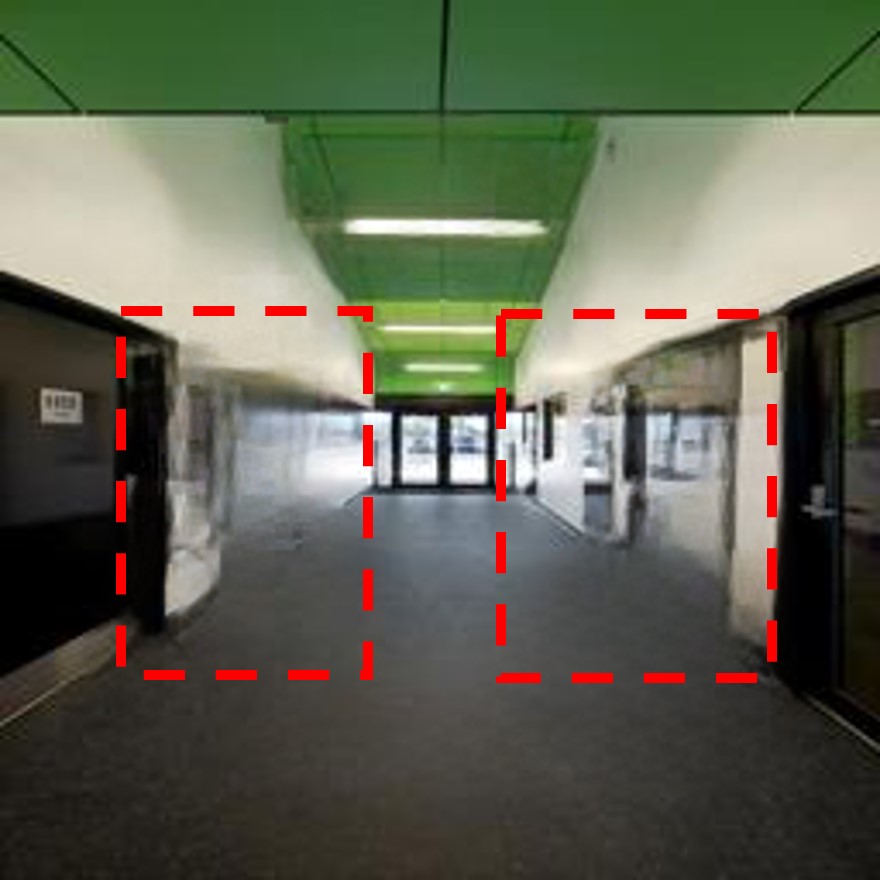}
    \includegraphics[width=0.1\textwidth]{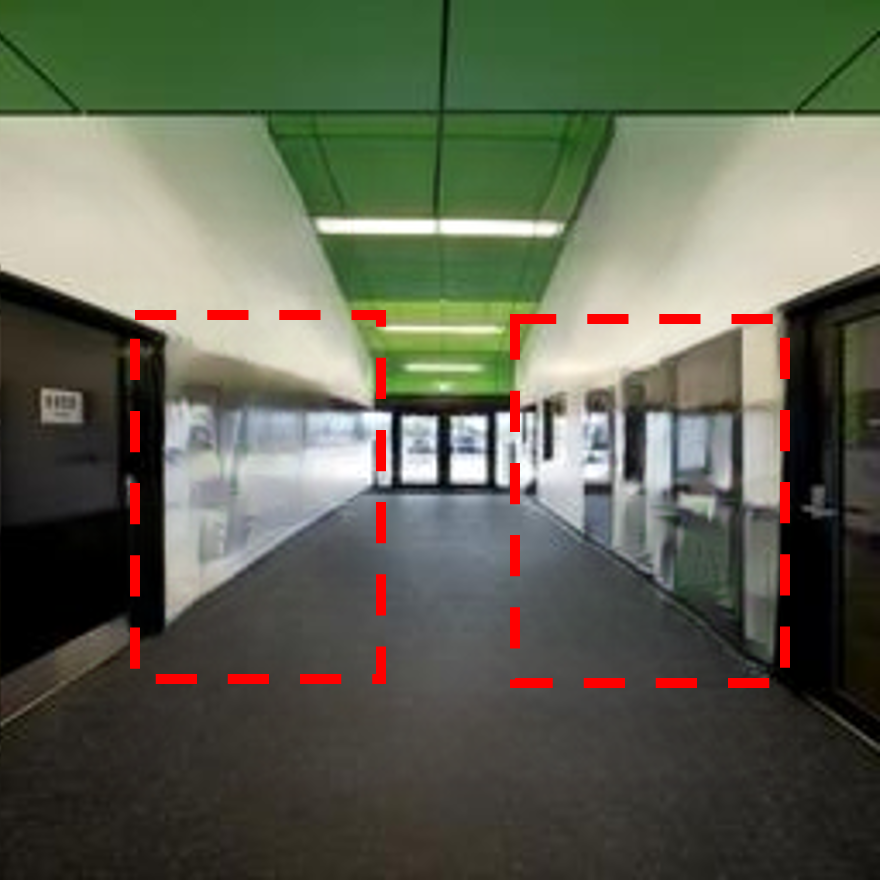}
    \includegraphics[width=0.1\textwidth]{images/places/gt2.jpg}
    \\
    \includegraphics[width=0.1\textwidth]{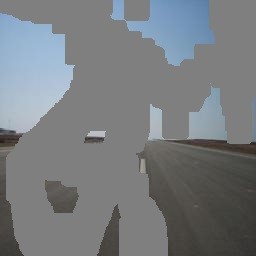}
    \includegraphics[width=0.1\textwidth]{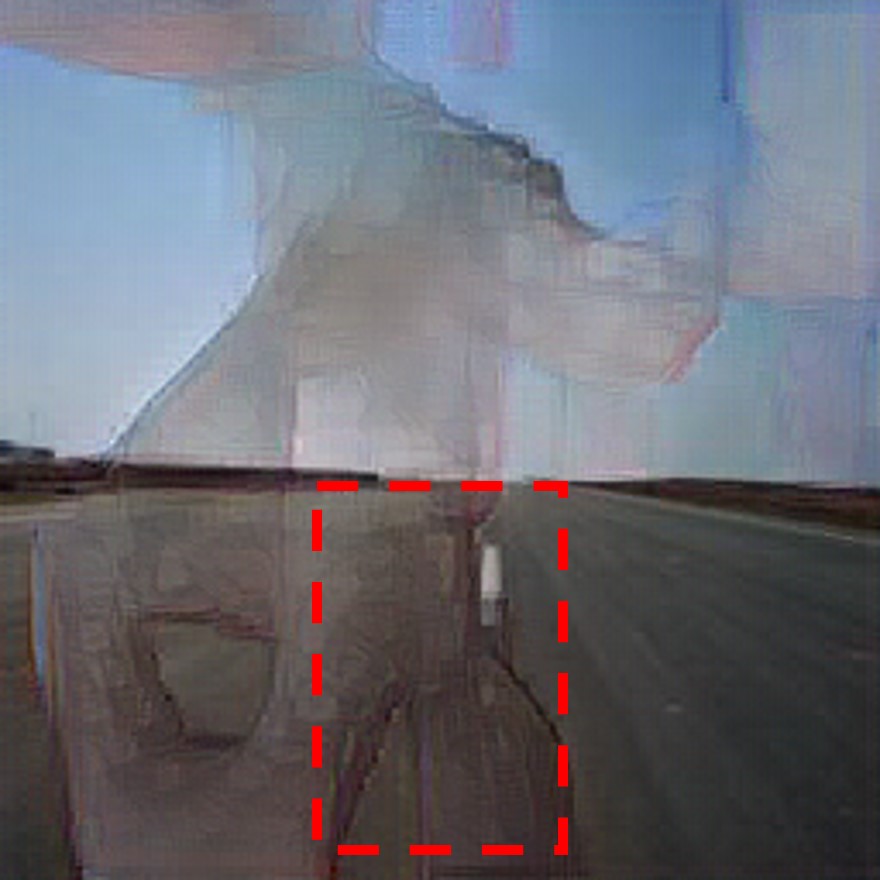}
    \includegraphics[width=0.1\textwidth]{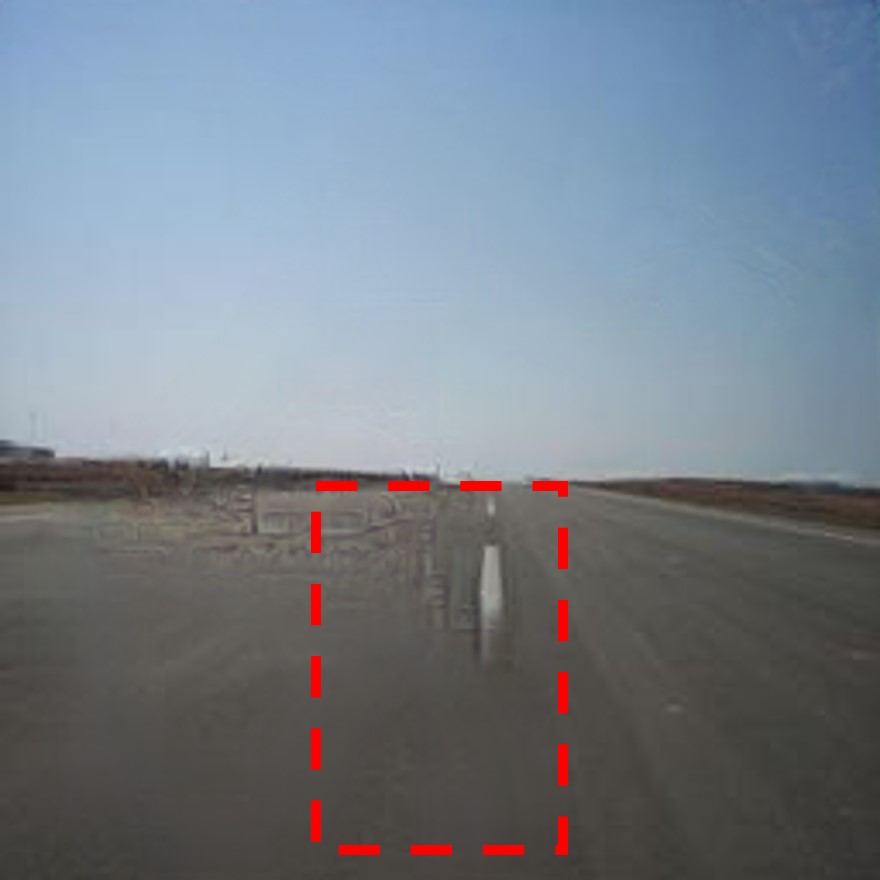}
    \includegraphics[width=0.1\textwidth]{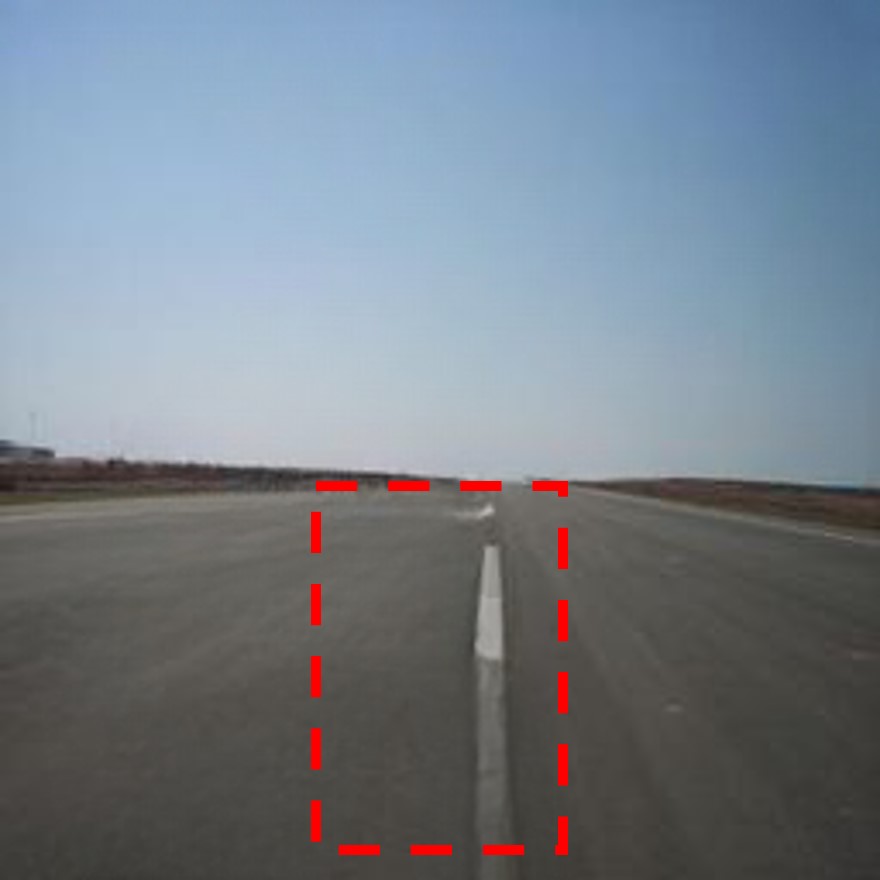}
    \includegraphics[width=0.1\textwidth]{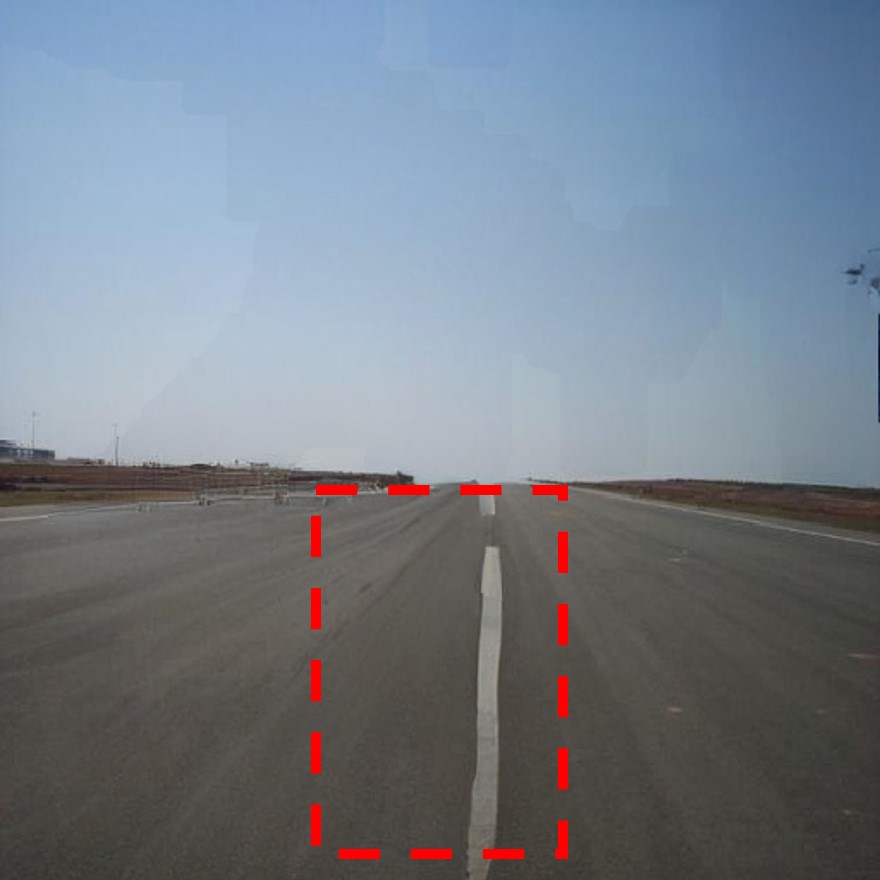}
    \includegraphics[width=0.1\textwidth]{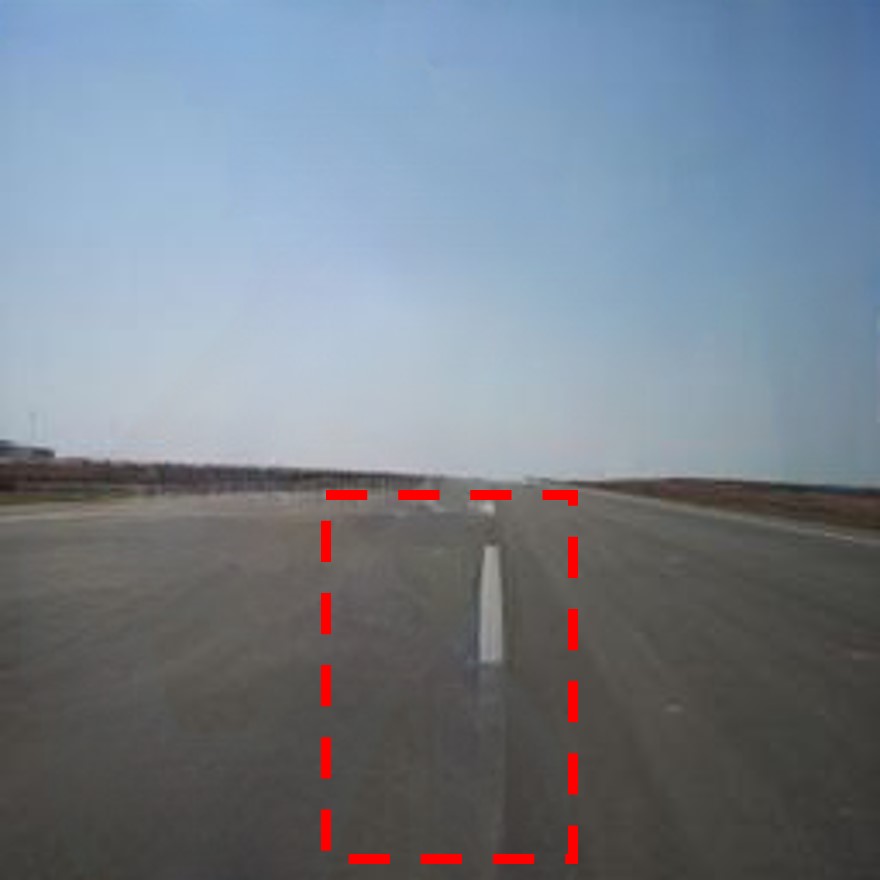}
    \includegraphics[width=0.1\textwidth]{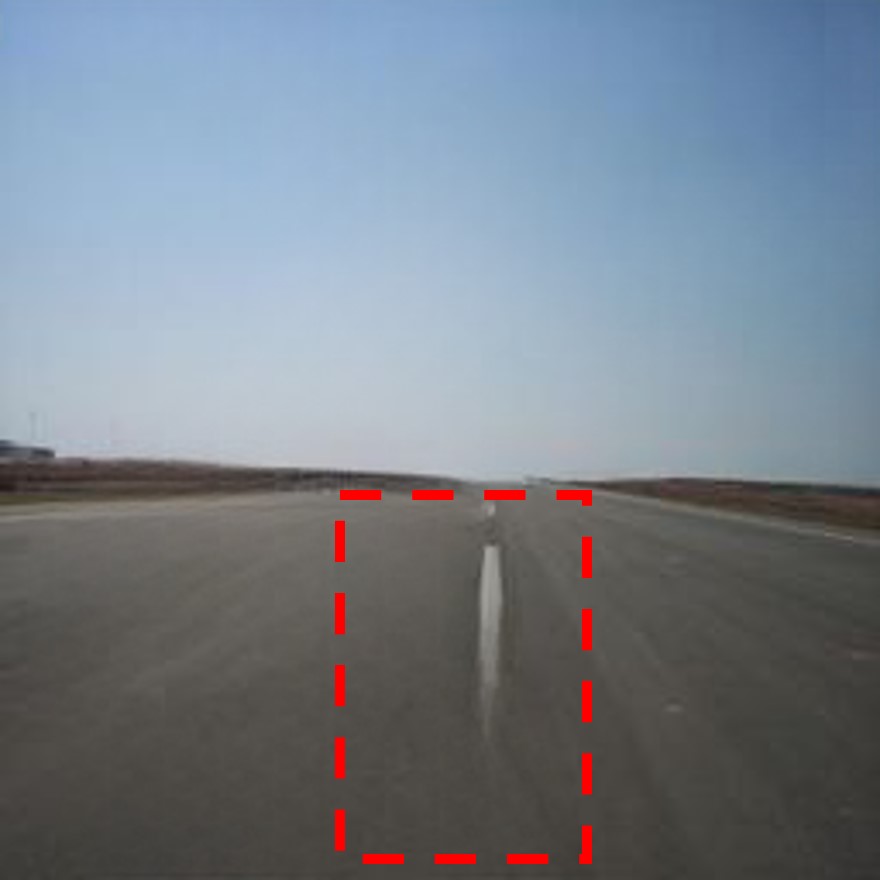}
    \includegraphics[width=0.1\textwidth]{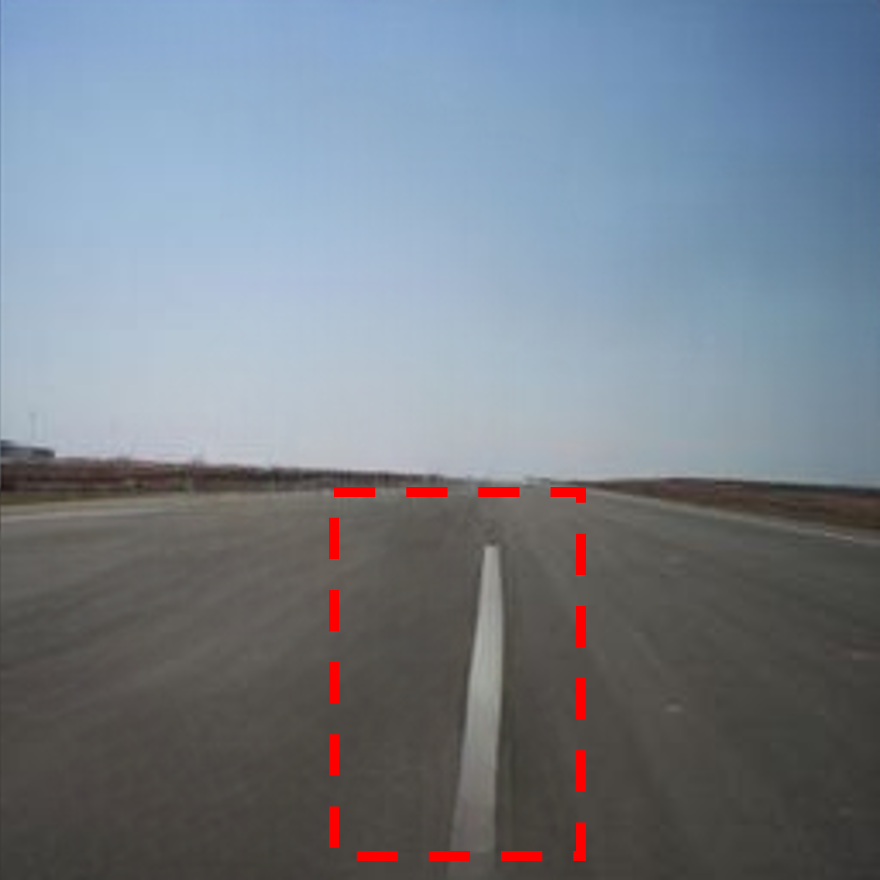}
    \includegraphics[width=0.1\textwidth]{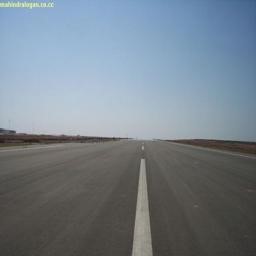}
    \\
    \caption{Visual comparisons at $(256 \times 256)$ resolution against the state-of-the-art methods on CelebA-HQ~\cite{karras2017progressive} (first two rows) and Places2~\cite{zhou2017places} (last two rows). } 
    \label{fig:comparison}
\end{figure*}

\noindent \textbf{Quantitative Comparison.}
For a fair comparison, we employ the officially released models and test them with the same test sets and masks. As shown in table~\ref{tab:long}, our \name{} outperforms in all metrics across different mask ratios. Especially on CelebA-HQ, at the increasing mask ratios, \name{} improves PSNR by 2.0\%, 2.3\% and 1.8\% respectively, and decreases LPIPS by 12.3\%, 11.6\% and 10.5\% respectively. Moreover, for Places2, our model demonstrated comparable performance to SOTAs such as MISF and LAMA. While our training utilized the Places2-Standard dataset with 1.8 million images, MISF and LAMA were trained on the Places2-Challenge dataset, which contains 8 million images. Despite using only 22.5\% of the images employed by these benchmarks, our model achieved comparable results, showcasing its robustness and efficiency.
\\
\\
\noindent \textbf{Qualitative Comparison.}
The qualitative comparisons are presented in Fig.~\ref{fig:comparison}. For human face samples, \name{} maintains consistency from the visible regions to the missing regions, such as effectively reconstructing elements like a missing hat. Additionally, \name{} renders features like eyes with improved fine-grained details, showcasing its strong capability in learning complex representations. 
In the Places2 dataset, \name{} effectively captures the spatial layouts in indoor environments and excels at maintaining the architectural integrity of the road surfaces. Such examples highlight our \name{} has superior spatial perceptions.
\\
\\
\noindent \textbf{Efficiency Comparison.}
Our \name{} efficiently reduces spatial dimensions within the transformer and leverages Mamba's inherent capabilities, both achieving linear complexity. This synergy optimizes operational simplicity while enhancing effectiveness. As shown in Tab.~\ref{tab:inf_time}, our model comprises 180 million parameters and achieves 110 ms to infer one image, making it suitable for real-time applications.

\begin{table*}[t]
\label{tab:ca}
	\centering
		\resizebox{\columnwidth}{!}{
		{\tabulinesep=0mm
			\begin{tabu}{@{\extracolsep{3pt}} c c c c c c c c c c | c c c c c | c c c c c@{}}
				\hline\hline
                    \multicolumn{1}{c}{\multirow{2}{*}{}} & 
				  \multicolumn{4}{c}{Components} & 
				\multicolumn{5}{c}{0.01\%-20\%} &
				\multicolumn{5}{c}{20\%-40\%} &
				\multicolumn{5}{c}{40\%-60\%}\T\B \\
					& MB & SRSA & GDFN~\cite{zamir2022restormer} & CBFN & PSNR$\uparrow$ & SSIM$\uparrow$ & L1$\downarrow$ & FID$\downarrow$ & LPIPS$\downarrow$ & PSNR$\uparrow$ & SSIM$\uparrow$ & L1$\downarrow$ & FID$\downarrow$ & LPIPS$\downarrow$ & PSNR$\uparrow$ & SSIM$\uparrow$ & L1$\downarrow$ & FID$\downarrow$  & LPIPS$\downarrow$\T\B\\
				\hline\hline
        \textbf{\textcolor{black}{(a)}} & & & & & 33.5812 & 0.9537 & 0.5385 & 1.4877 & 0.0513 & 25.8971 & 0.8527 & 1.9786 & 4.4025 & 0.1480 & 21.6134 & 0.7308 & 4.1254 & 8.1732 & 0.2464 \\
        \textbf{\textcolor{black}{(b)}} &\checkmark & &\checkmark & & 33.7640 & 0.9567 & 0.5244 & 1.4604 & 0.0425 & 26.1093 & 0.8736 & 1.8007 & 4.3721 & 0.1236 & 21.8573 & 0.7614 & 3.8649 & 8.0315 & 0.2197\\
        \textbf{\textcolor{black}{(c)}} & &\checkmark &\checkmark & & 33.7408 & 0.9598 & 0.5295 & 1.4181 & 0.0422 & 26.1274 & 0.8760 & 1.8092 & 4.3577 & 0.1196 & 21.8914 & 0.7682 & 3.8587 & 7.9974 & 0.2157 \\ 
        \textbf{\textcolor{black}{(d)}} &\checkmark &\checkmark &\checkmark & & 33.9042 & 0.9610 & 0.5129 & 1.4362 & 0.0419 & 26.2847 & 0.8768 & 1.7498 & 4.3115 & 0.1178 & 22.1377 & 0.7687 & 3.7679 & 7.9910 & 0.2067 \\
        \textbf{\textcolor{black}{(e)}} &\checkmark &\checkmark & &\checkmark & \textbf{34.1393} & \textbf{0.9618} & \textbf{0.5010} & \textbf{1.3896} & \textbf{0.0411} & \textbf{26.3231} & \textbf{0.8780} & \textbf{1.7043} & \textbf{4.2927} & \textbf{0.1170} & \textbf{22.1704} & \textbf{0.7699} & \textbf{3.6337} & \textbf{7.9905} & \textbf{0.2053}\B\\
        \hline
        \hline 
\end{tabu}
}
}
\vspace{0.5mm}
\caption{Ablation studies of each component. MB is the Mamba Block with positonal embedding. SRSA is the Spatial Reduced Self-Attention. GDFN is the feed-forward network in \cite{zamir2022restormer}. CBFN is the Context Broadcasting Feed-forward Network. Our \name{} corresponds to configuration (e).}
\vspace{-1mm}
\label{tab:abla}
\end{table*}
\begin{table}[htbp]
    \centering
    \begin{adjustbox}{max width=\textwidth}
    \begin{tabular}{l|ccccccccccc}
    \hline\hline
    Model & Wavefill~\cite{yu2021wavefill} &  WNet~\cite{zhang2022w} & MISF~\cite{li2022misf} & MAT~\cite{li2022mat} & LAMA~\cite{suvorov2022resolution} & CMT~\cite{ko2023continuously} & SD~\cite{Rombach_2022_CVPR} & LDM~\cite{Rombach_2022_CVPR} & Repaint~\cite{Lugmayr_2022_CVPR} & Ours \\
    \hline
    Param $\times 10^6$ & 49 & 46 & 26 & 62 & 51 & 143 & 860 & 387 & 552 & 180 \\
    Infer. Time (ms) & 70 & 35 & 10 & 70 & 25 & 60 & 880 & 6000 & 250000 & 110 \\
    \hline\hline
    \end{tabular}
    \end{adjustbox}
    \vspace{0.5mm}
    \caption{Comparison of parameter count and inference time from the evaluations conducted on $256\times256$ images.}
    \label{tab:inf_time}
\end{table}
\begin{figure*}[t] \centering

    \makebox[0.32\textwidth]{GT}
    \makebox[0.32\textwidth]{Masked Input}
    \makebox[0.32\textwidth]{Output}
    \\
    \begin{tabular}{l} 
        \includegraphics[width=0.32\textwidth]{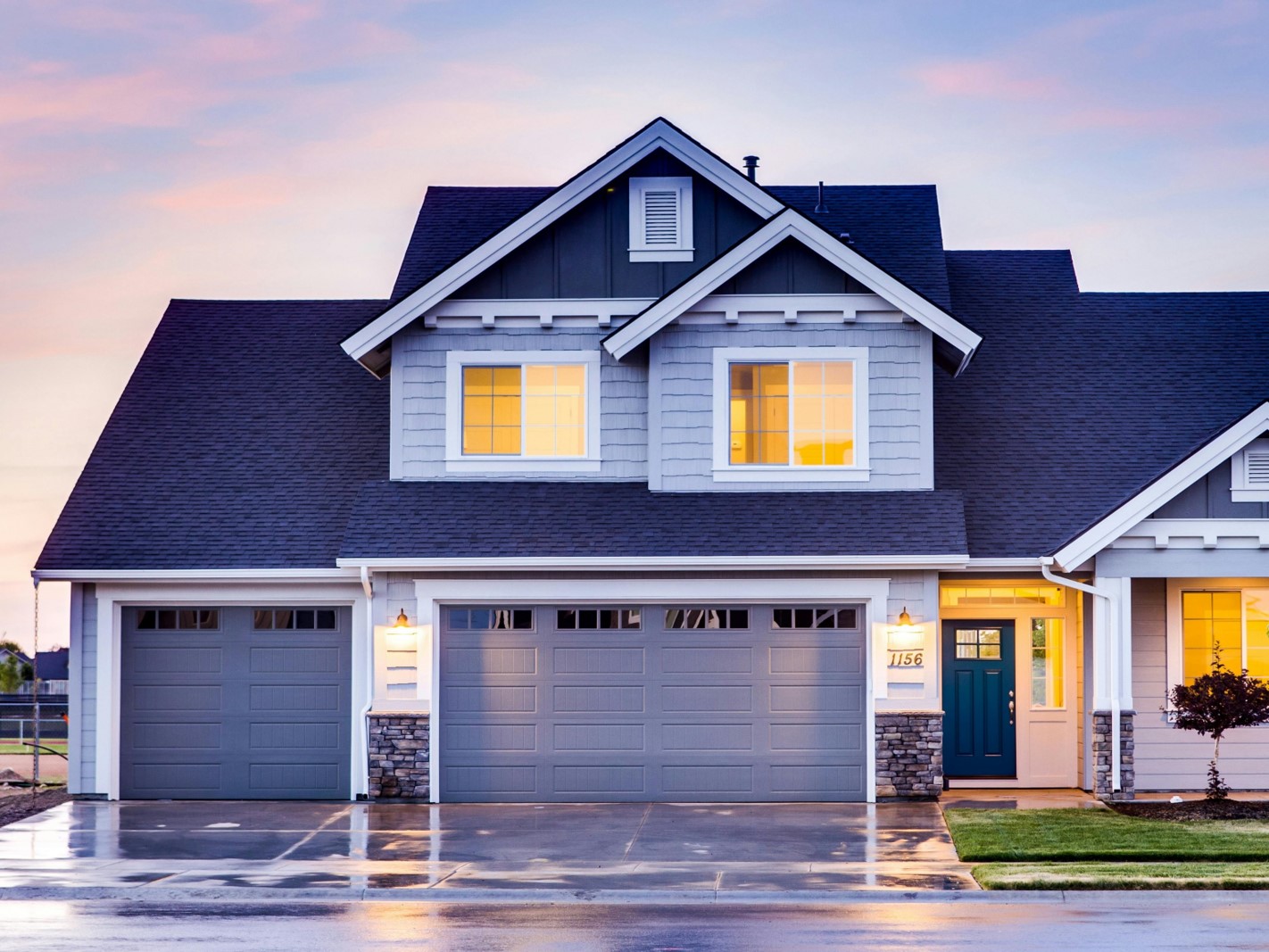}
        \includegraphics[width=0.32\textwidth]{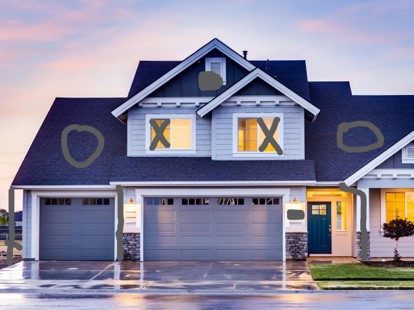}
        \includegraphics[width=0.32\textwidth]{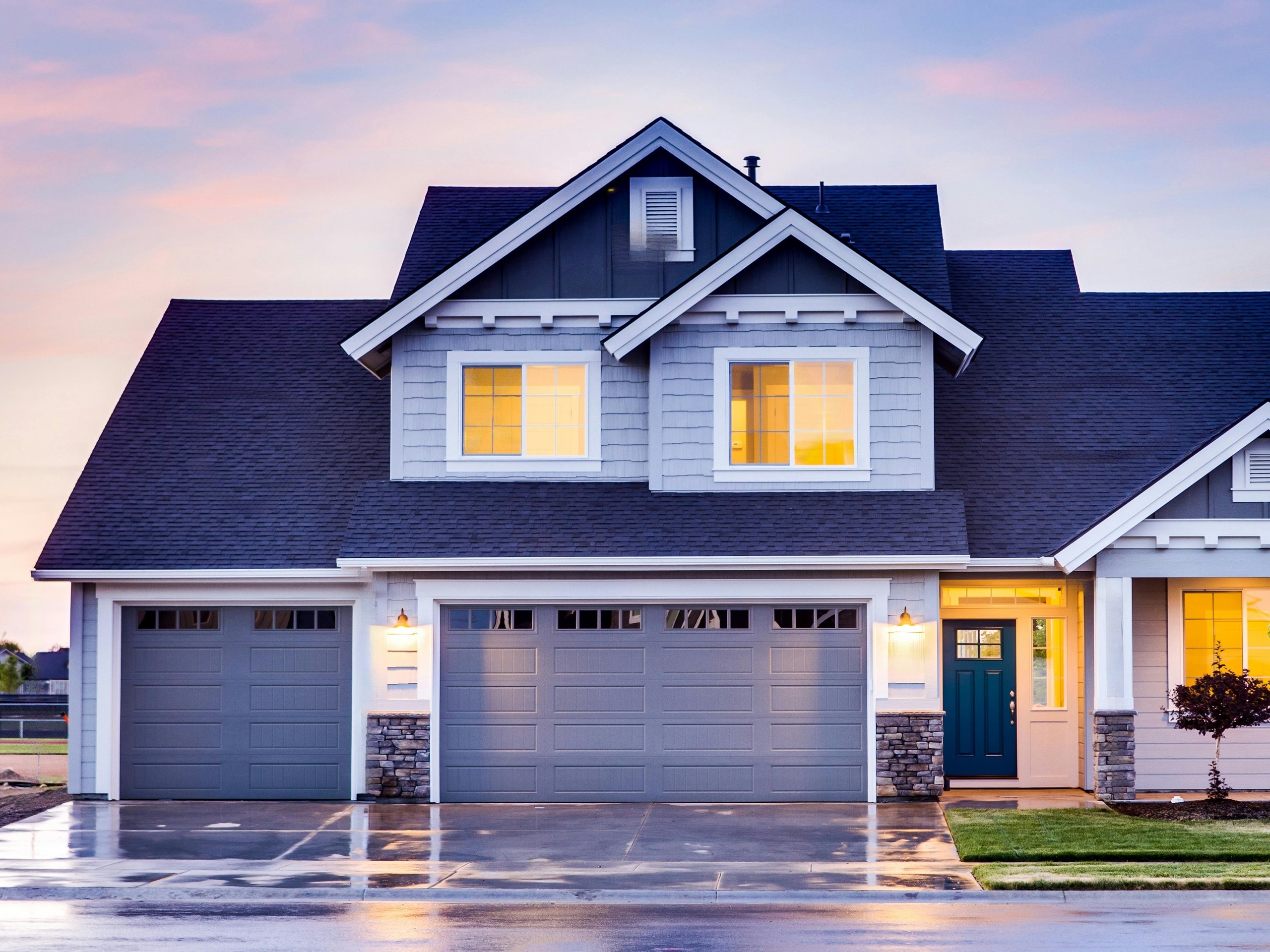} \\
    \end{tabular}
    \vspace{0.5em}
    \caption{Illustration of the application on real-world high-resolution images with resolution of $2560 \times 1920$.} 
    \label{fig:large}
\end{figure*}

\begin{figure*}[t] \centering

    \makebox[0.32\textwidth]{GT}
    \makebox[0.32\textwidth]{Masked Input}
    \makebox[0.32\textwidth]{Output}
    \\
    \begin{tabular}{l} 
        \includegraphics[width=0.32\textwidth]{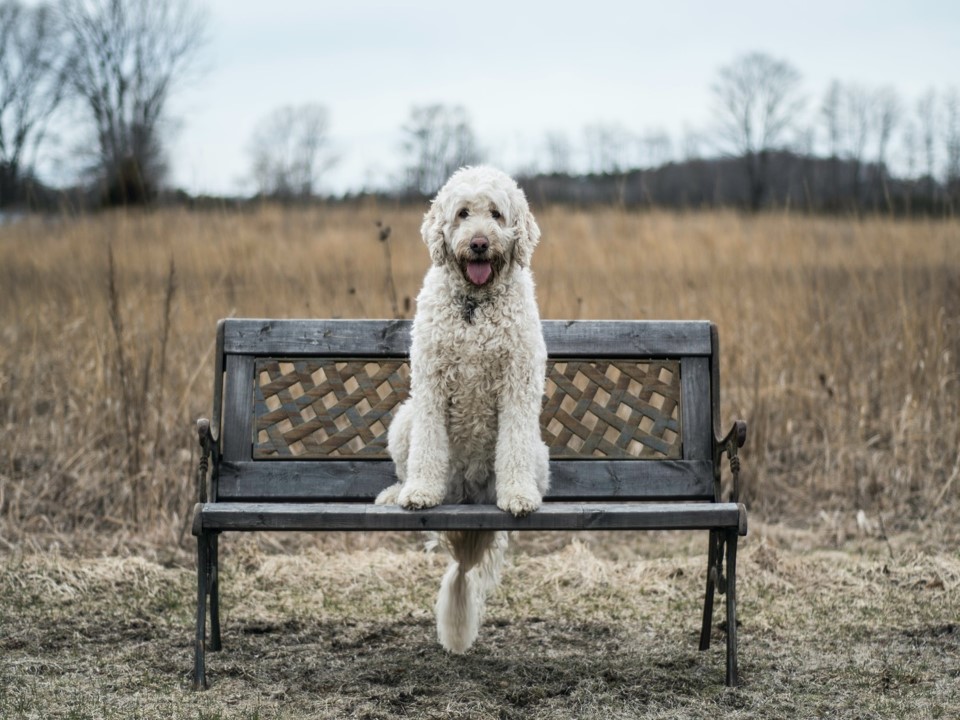}
        \includegraphics[width=0.32\textwidth]{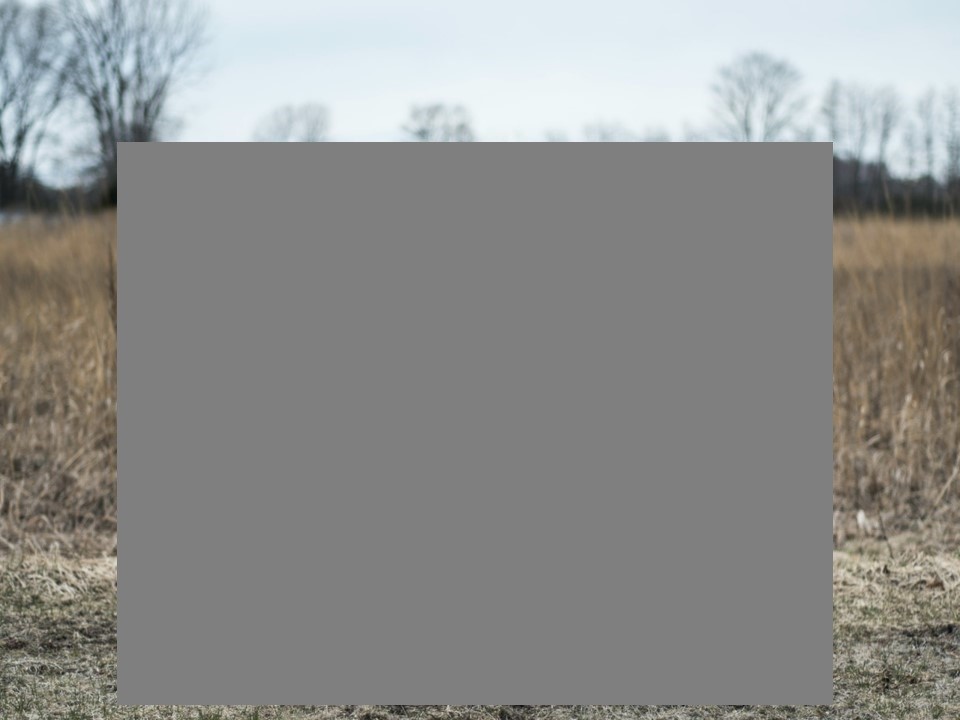}
        \includegraphics[width=0.32\textwidth]{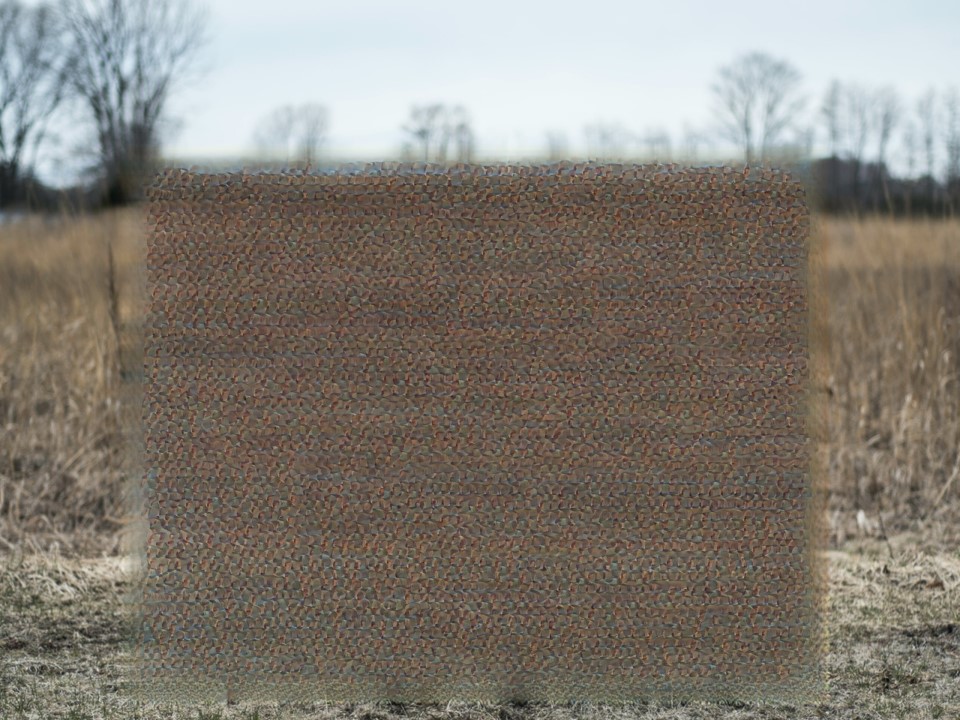} \\
    \end{tabular}
    \caption{Illustration of a fail case with a much larger mask ratio.} 
    \label{fig:fail}
\end{figure*}

\subsection{Ablation Study}
In our comprehensive ablation study conducted on CelebA-HQ, we incrementally enhance the baseline U-Net shape model, observing significant performance improvements with the integration of each component. Results are shown in the Tab.~\ref{tab:abla}. The addition of the Mamba Block in configuration (b) and the addition of self-attention in configuration (c) both demonstrated improvement across all evaluated metrics compared to the baseline (a). Notably, self-attention improves in SSIM, suggesting its superior capability in capturing spatial interactions. Mamba showcases the superior in capturing pixel-level interactions, demonstrated by the better PSNR and L1 values. The simultaneous use of Mamba and self-attention in configuration (d) lead to further improvements, indicating that these components effectively complement each other and contribute to a robust model. Configuration (e) is our final model, where we optimize GDFN to CBFN. The overall metrics are further improved. All ablation experiments are trained for 30K iterations. In addition, we followed ~\cite{cui2022selective} to build a light \name{} with a halved parameter for an efficient evaluation.


\subsection{Application: High-Resolution Image Inpainting}
Our \name{} is designed with linear computational complexity, enabling it to effectively handle high-resolution image inpainting tasks. We directly apply our model, pre-trained on the Places2-standard dataset, to real-world high-resolution images, demonstrating its capability. The example is illustrated in Fig.~\ref{fig:large}.

\subsection{Limitation and Discussion}
\noindent \textbf{Fail Cases.}
Our model is trained using a wildly used irregular mask dataset~\cite{liu2018partialinpainting}, where the largest mask ratio is 60\%, which restricts the ability to effectively handle images with much larger missing regions., particularly when the missing regions are concentrated in large shapes, such as very large rectangular masks (as shown in Fig.~\ref{fig:fail}).

\noindent \textbf{Future Work.}
In this work, our goal is to develop an inpainting model capable of reconstructing high-quality images, emphasizing fine-grained details and contextually plausible structures. Moving forward, our next objective is to enhance its capability by integrating multimodal foundational models, such as CLIP, to make the inpainting process controllable through text guidance, allowing users to influence the reconstruction results with descriptive input, while maintaining high quality.

\section{Conclusion}
In this paper, we introduce  \name{}, a hybrid model for image inpainting designed to reconstruct high-quality images with fine-grained details and spatial coherence. The proposed Hybrid Module effectively combines transformer and Mamba, leveraging the capacity of Mamba for capturing pixel-wise long-range interaction along with the spatial perception provided by the transformer. This integration enables \name{} to maintain linear computational complexity, which is particularly advantageous for handling high-resolution images. We validate \name{} on the widely-used CelebA-HQ and Places2 datasets, where it demonstrated superior or comparable performance to existing state-of-the-art methods.

\section{Acknowledgement}
This research is supported in part by the EPSRC NortHFutures project (ref: EP/X031012/1).


\bibliography{egbib}
\end{document}